\documentclass[10pt,twocolumn,letterpaper]{article}

\usepackage[pagenumbers]{iccv} % To force page numbers, e.g. for an arXiv version

\usepackage{booktabs}
\usepackage{colortbl} 
\usepackage[dvipsnames]{xcolor} % 使用 table 选项支持表格内着色
\usepackage{capt-of} 
\usepackage{graphicx}
\definecolor{lightblue}{RGB}{232, 244, 255}
\definecolor{paleRed}{RGB}{255,204,204}
\definecolor{palePurple}{RGB}{210,200,250}

\usepackage{multirow}
\usepackage[T1]{fontenc}
\definecolor{iccvblue}{rgb}{0.21,0.49,0.74}
\usepackage[pagebackref,breaklinks,colorlinks,allcolors=iccvblue]{hyperref}

\def\paperID{***} % *** Enter the Paper ID here
\def\confName{ICCV}
\def\confYear{2025}

\title{
    Less-to-More Generalization: \\ Unlocking More Controllability by In-Context Generation
}
\author{
Shaojin Wu\hspace{0.3cm}
Mengqi Huang$^{\ast}$\hspace{0.3cm}
Wenxu Wu\hspace{0.3cm}
Yufeng Cheng\hspace{0.3cm}
Fei Ding$^{\dag}$\hspace{0.3cm}
Qian He
\smallskip 
\\
Intelligent Creation Team, ByteDance
\smallskip 
\\
\tt\small\{wushaojin, huangmengqi.98, wuwenxu.01, chengyufeng.cb1, dingfei.212, heqian\}@bytedance.com \\
{\normalsize Project Page: \href{https://bytedance.github.io/UNO}{\bf\textcolor{orange}{https://bytedance.github.io/UNO}}}
}

\newcommand\blfootnote[1]{%
  \begingroup
  \renewcommand\thefootnote{}\footnote{#1}%
  \addtocounter{footnote}{-1}%
  \endgroup
}

\begin{document}
\twocolumn[{
\renewcommand\twocolumn[1][]{#1}
\maketitle
\begin{center}
    \captionsetup{type=figure}
    \vspace{-0.6cm}
    \includegraphics[width=0.95\textwidth]{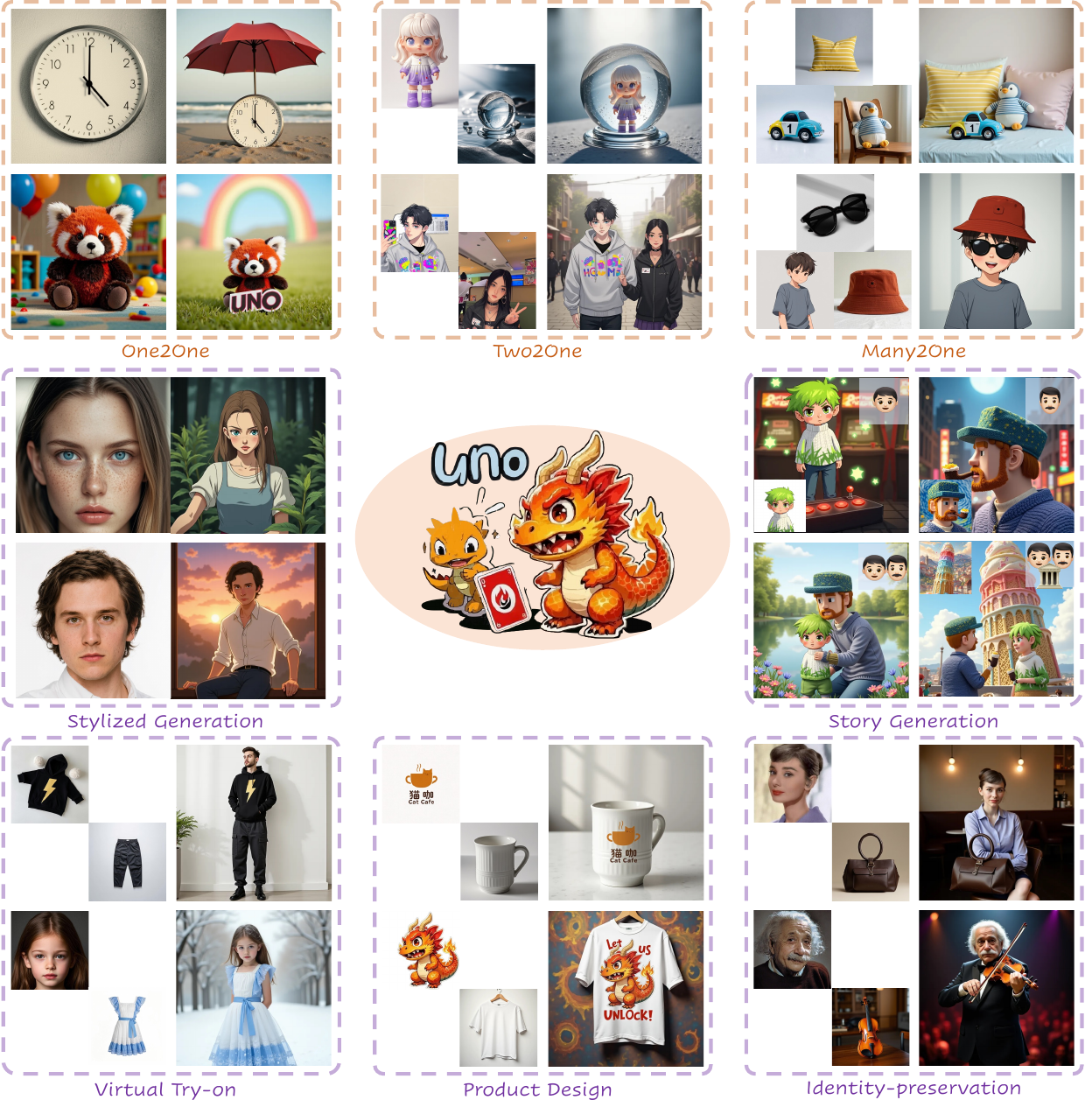}
    \vspace{-0.2cm}
    \captionof{figure}{Our UNO evolves as an universal customization from single to multi-subject.}
    \label{teaser}
\end{center}
}]

\blfootnote{$\ast$Corresponding author}
\blfootnote{$\dag$ Project lead\\}

\maketitle
\begin{abstract}
Although subject-driven generation has been extensively explored in image generation due to its wide applications, it still has challenges in data scalability and subject expansibility. For the first challenge, moving from curating single-subject datasets to multiple-subject ones and scaling them is particularly difficult. For the second, most recent methods center on single-subject generation, making it hard to apply when dealing with multi-subject scenarios. In this study, we propose a highly-consistent data synthesis pipeline to tackle this challenge. This pipeline harnesses the intrinsic in-context generation capabilities of diffusion transformers and generates high-consistency multi-subject paired data. Additionally, we introduce \textbf{UNO}, which consists of progressive cross-modal alignment and universal rotary position embedding. It is a multi-image conditioned subject-to-image model iteratively trained from a text-to-image model. Extensive experiments show that our method can achieve high consistency while ensuring controllability in both single-subject and multi-subject driven generation. Code and model: \href{https://github.com/bytedance/UNO}{\textcolor{orange}{https://github.com/bytedance/UNO}}.
\end{abstract}
\section{Introduction}
\label{sec:intro}

As the material medium for abstract linguistic semantics and spatial embodiments of concrete visual subjects, images constitute the foundational modality in intelligent content generation.
In recent years, customized image generation, which aims to create images that align with both the text semantics and the subjects in the reference images, has garnered significant interest across academic and industrial communities. This task unifies the flexibility of text control and the accuracy of visual controls, providing foundational infrastructure for diverse real-world applications ranging from film production to industrial design. As the field advances, the research challenge in customized image generation now centers on developing \emph{a stable and scalable paradigm for unlocking more controllability}, \ie, continuously increasing the amount of visual subject control without compromising the original text controllability.

Data, while serving as the foundation for training generative models, has long been the bottleneck in customized generation. 
An ideal model should be capable of generating visual subjects in diverse poses, locations, sizes, and other attributes based on text prompts. 
This necessitates data that encompasses multi-perspective subject variations, a requirement hindered by the impracticality of acquiring such comprehensive \emph{real paired datasets}.

Existing customized generation methods can be categorized into two streams based on \emph{how they utilize the data to design the model}, \ie, few-data fine-tuning and large-data training stream. Early few-data fine-tuning approaches \cite{ruiz2023dreambooth, gal2022image} primarily employ per-subject optimization through model fine-tuning \cite{ruiz2023dreambooth, kumari2023multi} or textual inversion \cite{gal2022image}, incurring substantial computational overhead and time-consuming during inference, which hinders real-world deployment. 
Therefore, more recent researches focus on the latter stream, which train adapters or image encoders on a large set of visual subjects to achieve real-time customization. 
Their corresponding data used for training are either real images with diversity limitations (eg, restricted subject variations \cite{yu2023mvimgnet, ye2023ip}), or synthetic data with limited image quality (typically, $\leq 512\times512$) and narrow domain coverage. 
Therefore, these methods often exhibit a trade-off between subject similarity and text controllability \cite{ye2023ip}, or unstable generation \cite{huang2024realcustom}.
Essentially, the existing customized models are designed based on their corresponding available customized data, resulting in limited scalability due to their data bottleneck.

\begin{figure}[t!]
    \centering
    \vspace{10pt}
    \includegraphics[width=1\linewidth]{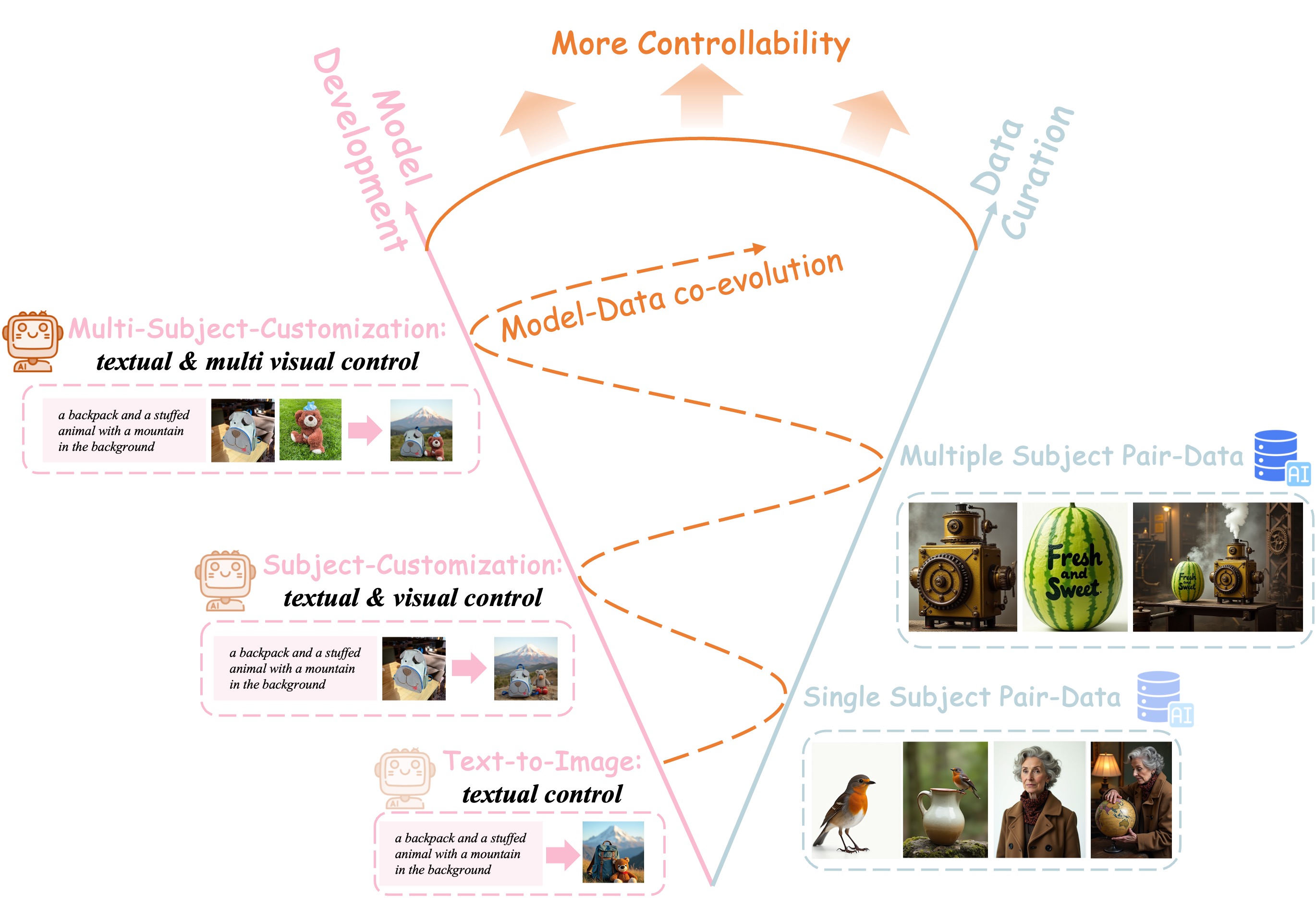}
    \vspace{10pt}
    \caption{The illustration of our motivation. We propose a novel \textbf{\emph{model-data co-evolution}} paradigm, where less-controllable preceding models systematically synthesize better customization data for successive more-controllable variants, enabling persistent co-evolution between enhanced model and enriched data.}
    \label{motivation}
\end{figure}

Diverging from the existing \emph{data-driven model design}, research on large language models (LLMs) demonstrates their capacity for strategic synthetic data generation toward model self-enhancement. 
Their bidirectional data-model knowledge transfer manifests through either high-performance models providing supervisory signals for weaker counterparts\cite{bai2022training, glaese2022improvingalignmentdialogueagents}, or conversely, less capable models could provide supervision to elicit higher capabilities leading to a stronger variant\cite{saunders2022self, burns2023weak, ouyang2022training}. 
Inspired by this LLMs' synthetic data-driven self-improvement, this study proposes that achieving \textbf{\emph{stable and scalable customized generation}} necessitates an analogous \textbf{\emph{model-data co-evolution}} paradigm, where less-controllable preceding customized models systematically synthesize better customization data for successive more-controllable variants, enabling persistent co-evolution between enhanced customized models and enriched customization data, as illustrated in Figure. \ref{motivation}.

Technically, to achieve the model-data co-evolution, this study addresses two fundamental challenges, \ie, 
(1) how to establish a \emph{systematic synthetic data curation framework} that reliably harnesses knowledge distillation from less-controllable models; and 
(2) how to develop a \emph{generalized customization model framework} capable of hierarchical controllability adaptation, ensuring seamless scalability across varying degrees of controllability.
To be specific, as for the \emph{synthetic data curation framework}, we introduce a progressive synthesis pipeline that transitions from single-subject to multi-subject in-context generation, combined with a multi-stage filtration mechanism to curate unprecedented high-resolution, high-quality paired customization data through fine-grained ensembled filtering.
As for the \emph{customization model framework}, we develop \textbf{UNO} to fully \textbf{un}l\textbf{o}ck the multi-condition contextual capabilities of Diffusion Transformers (DiT) through iterative simple-to-hard training, preserving the base architecture's scalability with minimal modifications. Moreover, we propose \textbf{un}iversal R\textbf{o}tary \textbf{P}osition \textbf{E}mbedding (UnoPE) to effectively equip UNO with the capability of mitigating the attribute confusion issue when scaling visual subject controls.

Our contributions are summarized as:

\textbf{Conceptual Contribution.} We identify that current data-driven approaches in customized model design inherently suffer from scalability constraints rooted in fundamental data bottlenecks. To address this limitation, we pioneer a model-data co-evolution paradigm that achieves enhanced controllability while enabling stable and scalable customized generation.

\textbf{Technical Contribution.} (1) We develop a systematic framework for synthetic data curation that produces high-fidelity, high-resolution paired customization datasets through progressive in-context synthesis and multi-stage filtering.
(2) We propose UNO, a universal customization architecture that enables seamless scalability across multi-condition control through minimal yet effective modification of DiT.

\textbf{Experimental Contribution.}We conduct extensive experiments on DreamBench~\cite{ruiz2023dreambooth} and multi-subject driven generation benchmarks. Our UNO achieve the highest DINO and CLIP-I scores among these two tasks. This demonstrates its strong subject similarity and text controllability, showcasing its capability to deliver state-of-the-art (SOTA) results.

\section{Related Work}
\label{sec:Related Work}

\subsection{Text-to-image generation}

Recent years have witnessed explosive growth in text-to-image (T2I) models~\cite{reed2016generative, ramesh2021zero, ding2021cogview, yu2022scaling, rombach2022high, podell2024sdxl, esser2024scaling, blackforestlabs_flux, xiao2024omnigen}.
Apart from some work that chooses the GAN or autoregressive paradigm, most of current text-to-image work chose the denoising diffuison~\cite{sohl2015deep, ho2020denoising} as their image generation framework.
Early exploratory work~\cite{nichol2021glide, ramesh2022hierarchical, saharia2022photorealistic} have validated the feasibility of using diffusion models for text-to-image generation and demonstrated their superior performance compared to other methods.
The efficiency, quality, and capacity for T2I diffusion models are keeping improved in the following work.
LDM~\cite{rombach2022high} suggests training the diffusion model in latent space significantly improves the efficiency and output resolution, which become the default choice for many subsequent works such as Stable Diffusion series~\cite{podell2024sdxl, esser2024scaling}, Imagen3~\cite{baldridge2024imagen} and Flux~\cite{blackforestlabs_flux}.
Recent work~\cite{peebles2023scalable, esser2024scaling, blackforestlabs_flux} replaces the unet~\cite{ronneberger2015u} to transformer and shows the impressive quality and scalability of the transformer backbone.

\subsection{Subject-driven generation}

Subject-driven generation has been widely studied in the context of diffusion models.
Dreambooth~\cite{ruiz2023dreambooth}, textual inversion~\cite{gal2022image} and LoRA~\cite{hu2021lora} introduce the subject-driven generation capability by introducing lightweight new parameters and perform parameter efficiency tuning for each subject.
The major drawback of those methods is the cumbersome fine-tuning process for each new-coming subject. 
IP-adapter~\cite{ye2023ip}, BLIP Diffusion~\cite{li2023blip} use an extra image encoder and new layers to encode the reference image of the subject and inject it into the diffusion model, achieving a subject-driven generation without further finetuning for a new concept.
For DiT, IC LoRA~\cite{huang2024context} and Ominicontrol~\cite{tan2024ominicontrol} have explored the inherent image reference capability in the transformer, pointing that the DiT itself can be used as the image encoder for the subject reference.
Many further works follow this reference image injection approach and have made improvements in various aspects like facial identity~\cite{wang2024instantid, guo2024pulid}, image-text joint controlability~\cite{huang2024realcustom}, multiple reference subject support~\cite{wang2025msdiffusion, mao2024realcustom++}.
Despite these advances, the aforementioned work heavily relies on paired images, which are hard to collect, especially for multi-subject scenes.

\begin{figure}[t]
\centering
\includegraphics[scale=0.36]{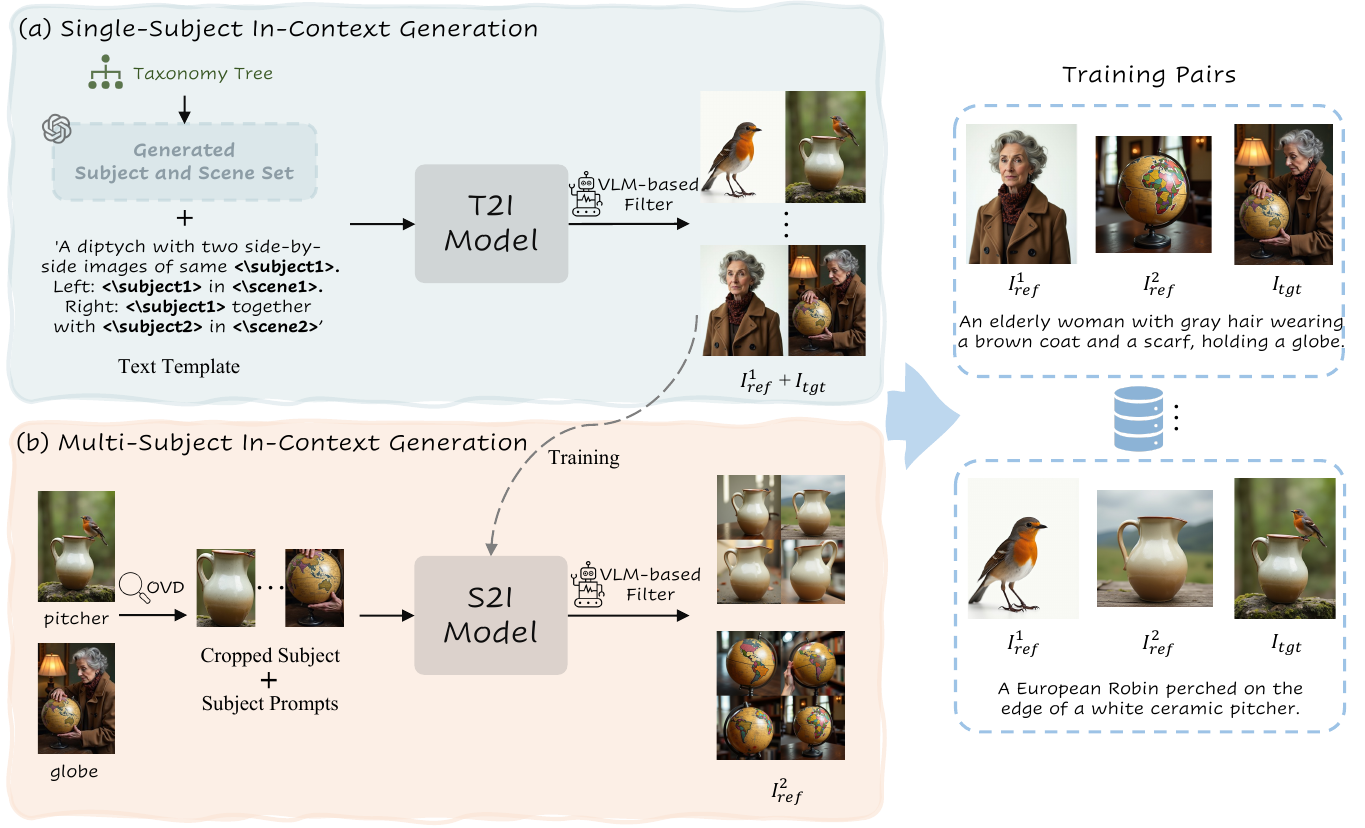}
    \caption{Illustration of our proposed synthetic data curation framework based on in-context data generation.}
    \label{fig2}
\end{figure}

\begin{figure*}[ht]
\centering
\includegraphics[scale=0.64]{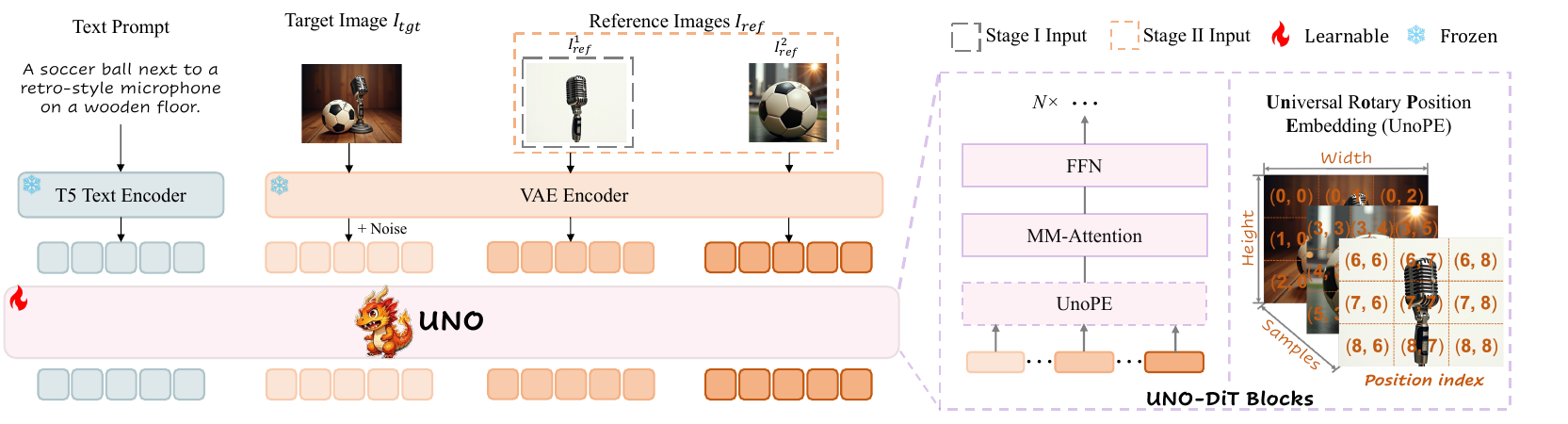}
    \caption{Illustration of the training framework of \textbf{UNO}. It introduces two pivotal enhancements to the model: progressive cross-modal alignment and universal rotary position embedding(UnoPE). The progressive cross-modal alignment is divided into two stages. In the  Stage $\mathrm{I}$, we use single-subject in-context generated data to finetune the pretrained T2I model into an S2I model. In the Stage $\mathrm{II}$, we continue training on generated multiple-subject data pairs. The UnoPE can effectively equip UNO with the capability of mitigating the attribute confusion issue when scaling visual subject controls.}
    \label{fig3}
\end{figure*}

% \begin{figure*}[ht]
% \centering
% \includegraphics[scale=0.90]{figs/score_step_rank.pdf}
%     \caption{TODO}
%     \label{fig5:score_step_thre}
% \end{figure*}

\section{Methodology}
\label{sec:Methodology}

This section introduces our proposed model-data co-evolution paradigm, encompassing the systematic synthetic data curation framework detailed in \cref{3.2} and the generalized customization model framework (\ie, UNO) expounded upon in \cref{3.3}. 
Specifically, the foundational work of DiT~\cite{peebles2023scalable} is elucidated in \cref{3.1}. 
\Cref{3.2} provides an in-depth exploration of the construction of our innovative subject-consistent dataset, comprising meticulously curated single-subject and multi-subject image pairs. 
Furthermore, \cref{3.3} outlines our methodology for transforming a Text-toImage (T2I) DiT model into a Subject-to-Image (S2I) model, showcasing its contextual generation capabilities. This adaptation involves an iterative training framework designed to facilitate multi-image perception, textual comprehension, and condition generation conducive to subject-driven synthesis.

% how we adapt a Text-to-Image (T2I) Diffusion Transformer (DiT~\cite{peebles2023scalable}) model into a Subject-to-Image (S2I) model with its in-context generation capabilities. Our methodology comprises three key components: First, we establish the theoretical foundation of diffusion transformers in \cref{3.1}. Subsequently, \cref{3.2} elaborates on the construction of our novel subject-consistent dataset containing carefully curated single-subject and multi-subject image pairs. Finally, \cref{3.3} delineates the iterative training framework for developing a DiT-based architecture capable of multi-image perception, textual understanding, and condition generation for subject-driven synthesis.
\subsection{Preliminary}
\label{3.1}
The original DiT architecture focuses solely on class-conditional image generation. It departs from the commonly used U-Net backbone, instead employing full transformer layers that operate on latent patches. More recently, image generators such as Stable Diffusion 3~\cite{esser2024scaling} and FLUX.1~\cite{blackforestlabs_flux} are built upon MM-DiT, which incorporates a multi-modal attention mechanism and takes as input the concatenation of the embeddings of text and image inputs.

The multi-modal attention operation projects position-encoded tokens into query $\mathbf{Q}$, key $\mathbf{K}$, and value $\mathbf{V}$ representations, enabling attention computation across all tokens:
\begin{equation}
\operatorname{Attention}\left([z_t, c]\right)=\operatorname{softmax}\left(\frac{\mathbf{Q} \mathbf{K}^\top}{\sqrt{d}}\right)\mathbf{V},\label{eq1}
\end{equation}
where $Z=[z_t, c]$ denotes the concatenation of image and text tokens. This allows both representations to function within their own respective spaces while still taking the other into account.

\subsection{Synthetic Data Curation Framework}
\label{3.2}
% The scarcity of high-quality, subject-consistent datasets has long posed a challenge for subject-driven generation, severely limiting the scalability of model training. As shown in \cref{fig2}, to address this issue, we propose a high-resolution, highly-consistent data synthesis pipeline that leverages the inherent in-context generation capabilities of DiT-based models. By employing well-designed text inputs, DiT models are capable of generating subject-consistent grid results. Unlike previous methods, such as OminiControl~\cite{tan2024ominicontrol}, which generate single-subject consistent data at a resolution of $512\times512$, we develop a more general pipeline that transitions from single-subject to multi-subject data generation and directly produce three different high-resolution (\ie, $1024\times1024$, $1024\times768$, and $768\times1024$) image pairs, significantly expanding the application scope and diversity of the synthesized data.
The paucity of high-quality, subject-consistent datasets has long presented a formidable obstacle for subject-driven generation, severely constraining the scalability of model training. As depicted in \cref{fig2}, we introduce a high-resolution, highly-consistent data synthesis pipeline to tackle this challenge, capitalizing on the intrinsic in-context generation capabilities of DiT-based models. Through the utilization of meticulously crafted text inputs, DiT models exhibit the capacity to generate subject-consistent grid outcomes. In contrast to prior methodologies like OminiControl~\cite{tan2024ominicontrol}, which generate single-subject consistent data at a resolution of $512\times512$, our approach establishes a more comprehensive pipeline that progresses from single-subject to multi-subject data generation. This advancement enables the direct production of three distinct high-resolution image pairs (\textit{i.e.}, $1024\times1024$, $1024\times768$, and $768\times1024$), significantly broadening the application spectrum and diversity of the synthesized data.

% We highlight that the high quality of the synthesized data can notably accelerate the convergence of models. To validate this hypothesis, we further constructed a filter based on Vision-Language Model (VLM) to score the generated image pairs and conducted experiments using synthesis data at different score levels.

\begin{figure}[t]
\centering
\includegraphics[scale=0.41]{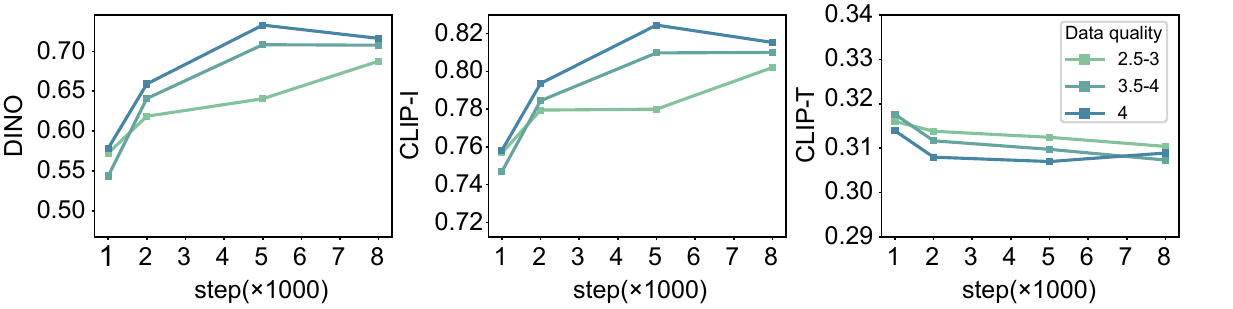}
    \caption{Model performance on Dreambench~\cite{ruiz2023dreambooth}. We conduct experiments under different quality score levels.}
    \label{fig4:score_step_thre}
\end{figure}

We emphasize that the superior quality of the synthesized data can significantly enhance model performance. To substantiate this assertion, we developed a filtering mechanism based on the Vision-Language Model (VLM) to evaluate the quality of the generated image pairs. Subsequently, we conducted experiments utilizing synthesized data across various quality score levels. As depicted in \cref{fig4:score_step_thre}, high-quality scores image pairs can significantly enhance the subject similarity of the results with higher DINO~\cite{oquab2023dinov2} and CLIP-I score, which verifies that our automated data curation framework can continuously supplement high-quality data and improve model performance.

\begin{figure*}[t]
\centering
\includegraphics[scale=0.72]{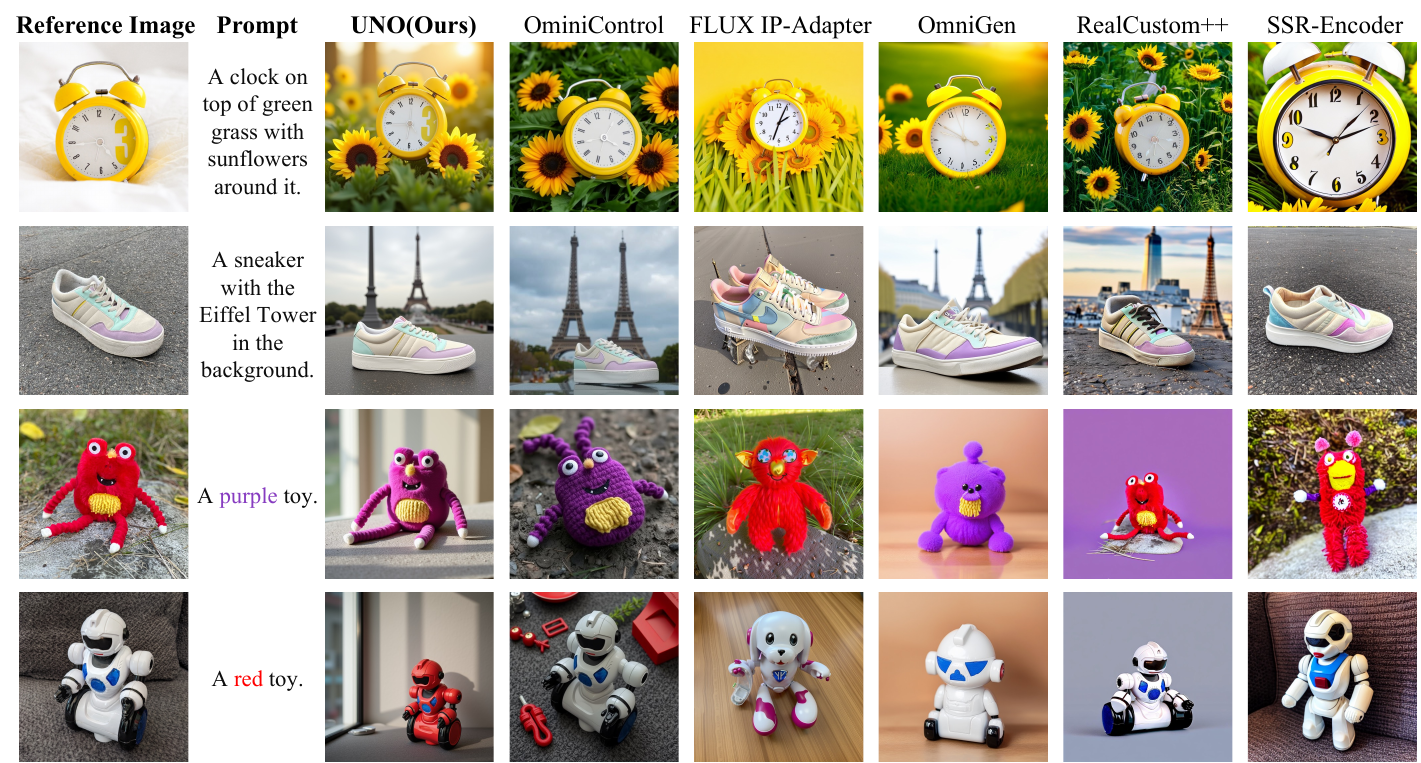}
    \caption{Qualitative comparison with different methods on single-subject driven generation.\newline}
    \label{fig4:single_ip}
\end{figure*}

\noindent \textbf{Single-Subject In-Context Generation.}
% To ensure dataset diversity, we first constructed a taxonomy tree that includes 365 general classes from Object365~\cite{shao2019objects365} and more fine-grained categories such as different ages, professions, and clothing styles. For each category, we utilize Large Language Model (LLM) to generate a vast number of subjects and diverse scenes. By combining these with predefined text templates, we can obtain millions of text prompts for T2I model to generate subject-consistent image pairs. 
In order to increase dataset diversity, we initially formulated a taxonomy tree comprising 365 overarching classes sourced from Object365~\cite{shao2019objects365}, alongside finer-grained categories encompassing distinctions in age, profession, and attire styles. Within each category, we leverage the capabilities of a Large Language Model (LLM) to generate an extensive array of subjects and varied settings. Through the amalgamation of these outputs with predefined text templates, we are able to derive millions of text prompts for the T2I model, facilitating the generation of subject-consistent image pairs.

Initially generated image pairs often encounter several issues, such as subject inconsistency and missing subjects. To efficiently filter the data, we first split the image pair into the reference image $I_{\text{ref}}^1$ and the target image $I_{\text{tgt}}$, then calculate the DINOv2~\cite{oquab2023dinov2}  between the two images. This method is effective in filtering out images with significantly lower consistency. Subsequently, VLM will be further employed to provide a score list evaluating different aspects (\ie, appearance, details, and attributes), which can be represented by the following equation:
\begin{equation}
\mathbf{S} = \operatorname{VLM}(I_{\text{ref}}, I_{\text{tgt}}, c_y)\in \mathbb{R}^{N \times 1},\label{eq2}
\end{equation}
\begin{equation}
\text{score} = \operatorname{Average}(\mathbf{S}), \label{eq3}
\end{equation}
where $c_y$ represents the input text to VLM, $N$ denotes the number of evaluated dimensions that are automatically generated by VLM, $\textbf{S}$ signifies the output which is parsed into a score list, and $\text{score}$ indicates the final consistency score of generated image pairs. 

\noindent \textbf{Multi-Subject In-Context Generation.}
The comprehensive dataset from the preceding phase will be utilized to train a S2I model, which is conditioned on both single-image and text inputs. Subsequently, this trained S2I model, along with the dataset, will be employed to generate multi-subject consistent data in the current stage. Illustrated in \cref{fig2}(b), we will initially employ an open-vocabulary detector (OVD) to identify subjects beyond those present in $I_{\text{ref}}^1$. The extracted cropped images and their corresponding subject prompts will then be input into our trained S2I model to derive new results for $I_{\text{ref}}^2$. Traditional approaches often encountered significant failure rates as models struggled to preserve the original subject's identity. However, models trained using our proposed in-context training methodology can effectively surmount this challenge, yielding highly consistent outcomes with ease. Further elaboration on this topic will be provided in the subsequent section.
% The full-mark data from the previous stage will be used to train an S2I model which is conditioned on single-image and text inputs. Then, both the data and this trained S2I model will be applied to generate multi-subject consistent data in this stage. As depicted in \cref{fig2}(b), we will initially utilize an open-vocabulary detector (OVD) to identify subjects outside of those in $I_{\text{ref}}^1$. The cropped images and their corresponding subject prompts are then fed into our trained S2I model to obtain new $I_{\text{ref}}^2$ results. Previous methods often had a high failure rate as models struggled to maintain the identity of the original subject. However, the model trained with our proposed in-context training method can effectively overcome this challenge and easily produce highly consistent results. We will elaborate on this in the next section. 

Some may question the necessity of generating new data, suggesting that the cropped part could be treated simply as $I_{\text{ref}}^2$. However, we contend that relying solely on cropped images as the training dataset may introduce copy-paste issues. This scenario arises when the model fails to adhere to the textual prompt and merely ``pastes" the input reference image onto the resulting image. Our proposed pipeline effectively mitigates this concern.

% Some may question the necessity of generating new data since we could simply treat the cropped part as $I_{\text{ref}}^2$. We argue that using just the cropped images as the training dataset may easily lead to copy-paste issues. This is when the model does not follow the text prompt and merely $pastes$ the input reference image onto the result image. Our proposed pipeline can effectively avoid this problem.

\begin{figure*}[ht]
\centering
\includegraphics[scale=0.725]{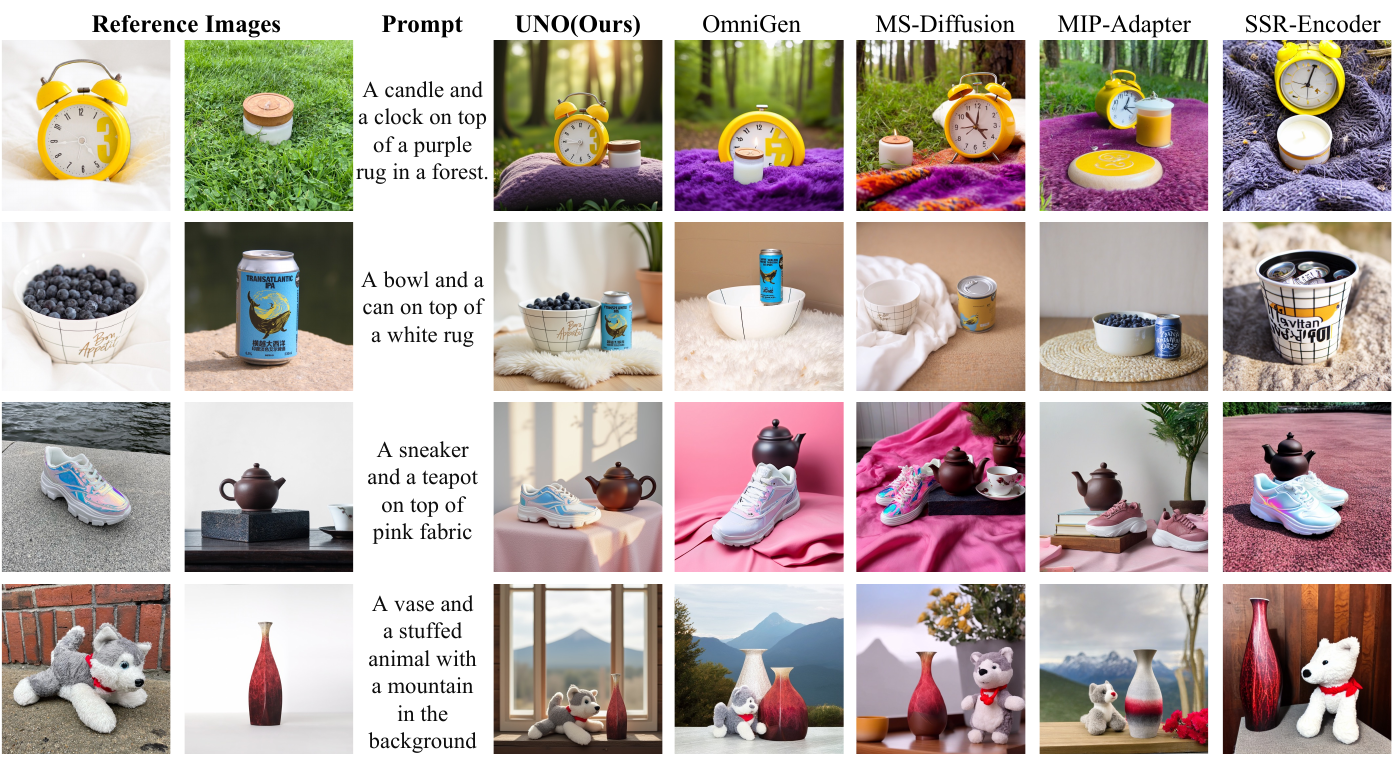}
    \caption{Qualitative comparison with different methods on multi-subject driven generation.\newline}
    \label{fig5:multi-ip}
\end{figure*}

\subsection{Customization Model Framework (UNO)}
\label{3.3}
In this section, we will provide a detailed explanation of how to iteratively train a multi-image conditioned S2I model from a DiT-based T2I model. It should be noted that all the training data we used originate from images generated by our in-context data generation method proposed in the previous chapter.

\noindent \textbf{Progressive cross-modal alignment.} Original T2I models gradually transform pure Gaussian noise into text-adherent images through an iterative denoising process. During this process, the VAE encoder $\mathcal{E}(\cdot)$ first encodes the target image $I_{\text{tgt}}$ into a noisy latent $z_t = \mathcal{E}(I_{\text{tgt}})$. $z_t$ is then concatenated with the encoded text token $c$, forming the input $z$ for the DiT model. This process can be formulated as:
\begin{equation}
z = \operatorname{Concatenate}(c, z_t), \label{eq4}
\end{equation}

% To incorporate multi-image conditions $I_{\text{ref}} = [I_{\text{ref}}^1, I_{\text{ref}}^2, \ldots, I_{\text{ref}}^N]$, we propose a less-to-more training paradigm. We consider the training phase with single-image conditions as a cross-modal alignment phase. Since the original input only contains text tokens and noisy latents, the newly added noise-free reference image tokens could potentially disrupt the original convergence distribution. This disruption may lead to training instability or sub-optimal results. Therefore, we adopt a simple-to-complex approach rather than directly allowing the model to receive multiple reference image inputs. As shown in \cref{fig3}, Stage $\mathrm{I}$ only uses a single image as the reference image. We use $z_1$ as the input multi-modal tokens for the DiT model:
To incorporate multi-image conditions $I_{\text{ref}} = [I_{\text{ref}}^1, I_{\text{ref}}^2, \ldots, I_{\text{ref}}^N]$, we introduce a progressive training paradigm that progresses from simpler to more complex scenarios. We view the training phase with single-image conditions as the initial phase for cross-modal alignment. Given that the original input comprises solely text tokens and noisy latents, the introduction of noise-free reference image tokens could potentially disrupt the original convergence distribution. Such disruption may result in training instability or suboptimal outcomes. Hence, we opt for a gradual complexity approach rather than directly exposing the model to multiple reference image inputs. In Stage $\mathrm{I}$, depicted in \cref{fig3}, only a single image serves as the reference image. We utilize $z_1$ as the input multi-modal tokens for the DiT model:
\begin{equation}
z_1 = \operatorname{Concatenate}(c, z_t, \mathcal{E}(I_{\text{ref}}^1)), \label{eq5}
\end{equation}
After Stage $\mathrm{I}$ training, the model is capable of processing single-subject driven generation tasks. We then train the model with multi-image conditions to tackle more complex multi-subject driven generation scenarios. $z_2$ can be described as follows:
\begin{equation}
z_{\text{ref}}^i = \mathcal{E}(I_{\text{ref}}^i), \quad i=1,\ldots,N,\label{eq7}
\end{equation}
\begin{equation}
z_2 = \operatorname{Concatenate}(c, z_t, z_{\text{ref}}^1, z_{\text{ref}}^2, \ldots, z_{\text{ref}}^N), \label{eq8}
\end{equation}
where $N$ is set to $2$ in our paper. During Stage $\mathrm{I}$, the T2I model is trained to refer to the input reference image and prompt, with the goal of generating single subject-consistent results. Stage $\mathrm{II}$ is designed to enable the S2I model to refer to multiple input images and inject information into corresponding latent spaces. Through iterative training, the inherent in-context generation capability of T2I model is unlocked, eliciting more controllability from single text-to-image models.

\noindent \textbf{Universal rotary position embedding(UnoPE).} An important consideration for incorporating multi-image conditions into a DiT-based T2I model pertains to the aspect of position encoding. In the context of FLUX.1~\cite{blackforestlabs_flux}, the utilization of Rotary Position Embedding (RoPE) necessitates the assignment of position indices $(i, j)$ to both text and image tokens, thereby influencing the interaction among multimodal tokens. Within the original model architecture, text tokens are assigned a consistent position index of $(0, 0)$, while noisy image tokens are allocated position indices $(i, j)$ where $i \in [0, w - 1]$ and $j \in [0, h - 1]$. Here, $h$ and $w$ denote the height and width of the noisy latent, respectively.

% Another question to apply multi-image conditions to a DiT-based T2I model lies in the position encoding. In FLUX.1~\cite{blackforestlabs_flux}, the used Rotary Position Embedding(RoPE) requires both text and image tokens to be assigned a position index $(i, j)$, which affects the interaction between multi-modal tokens. In the original model architecture, text tokens maintain a fixed position index of $(0, 0)$, and noisy image tokens are assigned position indices $(i, j)$ where $i \in [0, w - 1]$ and $j \in [0, h - 1]$. Here, $h$ and $w$ correspond to the height and width of the noisy latent, respectively.

Our newly introduced image conditions reuse the same format to inherit the implicit position correspondence of the original model. However, we start from the maximum height and width of the noisy image tokens, as shown in \cref{fig3}, which begins with the diagonal position. The position index for the latent $z_{\text{ref}}^N$ is defined as:
\begin{equation}
    (i', j') = (i + w^{(N-1)}, j + h^{(N-1)}),
\end{equation}
where $ i \in [0, w^N) $, $ j \in [0, h^N) $, with $w^N$ and $h^N$ representing the width and height of the latent $z_{\text{ref}}^N$, respectively. Here, $i'$ and $j'$ are the adjusted position indices. To prevent the generated image from over-referencing the spatial structure of the reference image, we adjust the position indices within a certain range. In the scenario of multi-image conditions, different reference images inherently have a semantic gap. Our proposed UnoPE can further prevent the model from learning the original spatial distribution of reference images, thereby focusing on obtaining layout information from text features. This enables the model to improve its performance in subject similarity while maintaining good text controllability.

% dia pe single-ip+multi-ip, two column
%%
% \begin{table*}[t]
% \centering
% \resizebox*{0.7\textwidth}{!}{
% \begin{tabular}{lccccccc}
% \toprule
% \multirow{2}{*}{Method} & \multicolumn{3}{c}{Single-subject} & \multicolumn{3}{c}{Multi-subject} \\
% \cmidrule(lr){2-4} \cmidrule(lr){5-7}
%  & \textbf{DINO} $\uparrow$ & \textbf{CLIP-I} $\uparrow$ & \textbf{CLIP-T} $\uparrow$ & \textbf{DINO} $\uparrow$ & \textbf{CLIP-I} $\uparrow$ & \textbf{CLIP-T} $\uparrow$ \\\hline
% w/o shift PE & 0.470 & 0.722 & 0.308\\
% w/ w-shift PE & 0.717  & 0.813 & 0.304\\
% w/ h-shift PE & 0.678  & 0.797 & 0.308\\\hline
% \textbf{Ours} & \textbf{0.730}  & \textbf{0.821} & \textbf{0.309}\\
% \bottomrule
% \end{tabular}
% }
% \caption{Effect of in-context data generation. Model trained with multiple subject pair-data performs better on DreamBench~\cite{} compared to the model trained only with single subject pair-data, with both models trained for the same number of steps.}
% \label{tab3}
% \end{table*}

\section{Experiments}
\label{sec:Experiments}

\subsection{Experiments Setting}
\noindent \textbf{Implementation Details.}
To self-evolution our base DiT-based T2I model, we firstly take the FLUX.1 dev~\cite{blackforestlabs_flux} as the pretrained model. We train the model with a learning rate of $10^{-5}$ and a total batch size of $16$. For the progressive cross-modal alignment, we first train the model using single-subject pair-data for $5,000$ steps. Then, we continue training on multi-subject pair-data for another $5,000$ steps. Specifically, we generated $230k$ and $15k$ data pairs for these two stages respectively using the in-context data generation method mentioned above. We conduct the entire experiment on $8$ NVIDIA A100 GPUs and trained the model using a LoRA~\cite{hu2021lora} rank of $512$ throughout the training process.

\noindent \textbf{Comparative Methods.} As a tuning-free method, our model is capable of handling both single-subject and multi-subject driven generation. We compare it with some leading methods in these two tasks respectively, including Omnigen~\cite{xiao2024omnigen}, Ominicontrol~\cite{tan2024ominicontrol}, FLUX IPAdapter v2~\cite{flux-ipav2}, Ms-diffusion~\cite{wang2025msdiffusion}, MIP-Adapter~\cite{huang2024resolving}, RealCustom++~\cite{mao2024realcustom++}, and SSR-Encoder~\cite{zhang2024ssr}.

\noindent \textbf{Evaluation Metrics.}
Following previous works, we use standard automatic metrics to evaluate both subject similarity and text fidelity. Specifically, we employ cosine similarity measures between generated images and reference images within CLIP~\cite{radford2021learning} and DINO~\cite{oquab2023dinov2} spaces, referred to as CLIP-I and DINO scores, respectively, to assess subject similarity. Additionally, we calculate the cosine similarity between the prompt and the image CLIP embeddings (CLIP-T) to evaluate text fidelity. For single-subject driven generation, we measure all methods on DreamBench~\cite{ruiz2023dreambooth} for fairness. For multi-subject driven generation, we follow previous studies~\cite{ma2024subject, huang2024resolving} that involve $30$ different combinations of two subjects from DreamBench, including combinations of non-live and live objects. For each combination, we generate $6$ images per prompt using $25$ text prompts from DreamBench, resulting in $4,500$ image groups for all subjects.

\begin{table}[t]
\centering
\small
% \resizebox*{0.425\textwidth}{!}{
% \resizebox*{0.5\textwidth}{!}{
\begin{tabular}{lcccc}
\toprule
\textbf{Method} & \textbf{DINO} $\uparrow$ & \textbf{CLIP-I} $\uparrow$ & \textbf{CLIP-T} $\uparrow$ \\ \toprule
Oracle(reference images) & 0.774 & 0.885 & - \\\hline
Textual Inversion \cite{gal2022image} & 0.569 & 0.780 & 0.255\\
DreamBooth \cite{ruiz2023dreambooth} &0.668  & 0.803 & 0.305 \\
BLIP-Diffusion \cite{li2023blip} & 0.670 & 0.805 & 0.302\\\hline
ELITE \cite{wei2023elite}& 0.647  & 0.772 & 0.296\\
Re-Imagen \cite{chen2022re}& 0.600  & 0.740 & 0.270\\
BootPIG\cite{purushwalkam2024bootpig} & 0.674  & 0.797 & 0.311\\
SSR-Encoder\cite{zhang2024ssr} & 0.612  & \colorbox{palePurple}{0.821} & 0.308\\
RealCustom++ \cite{huang2024realcustom, mao2024realcustom++}& \colorbox{palePurple}{0.702}  & 0.794 & \colorbox{paleRed}{\textbf{0.318}}\\
OmniGen \cite{xiao2024omnigen}& 0.693  & 0.801 & \colorbox{palePurple}{0.315}\\
OminiControl \cite{tan2024ominicontrol}& 0.684  & 0.799 & 0.312\\
FLUX.1 IP-Adapter & 0.582  & 0.820 & 0.288\\
\textbf{UNO (Ours)} & \colorbox{paleRed}{\textbf{0.760}}  & \colorbox{paleRed}{\textbf{0.835}} & 0.304\\
\bottomrule
\end{tabular}
%}
\caption{\textbf{Quantitative results for single-subject driven generation on Dreambench.} We present the oracle results in the first row and compare both tuning methods and tuning-free methods. We highlight the \colorbox{paleRed}{best} and \colorbox{palePurple}{second-best} values for each metric.}
\label{tab1:single_ip}
\end{table}

\begin{table}[t]
\centering
\small
%\resizebox*{0.4\textwidth}{!}{
\begin{tabular}{lcccc}
\toprule
\textbf{Method} & \textbf{DINO} $\uparrow$ & \textbf{CLIP-I} $\uparrow$ & \textbf{CLIP-T} $\uparrow$ \\ \toprule
DreamBooth \cite{ruiz2023dreambooth} &0.430  & 0.695 & 0.308 \\
BLIP-Diffusion \cite{li2023blip} & 0.464 & 0.698 & 0.300\\\hline
Subject Diffusion \cite{ma2024subject}& 0.506  & 0.696 & 0.310\\
MIP-Adapter \cite{huang2024resolving}& 0.482  & \underline{0.726} & 0.311\\
MS-Diffusion \cite{wang2025msdiffusion}& \colorbox{palePurple}{0.525}  & \colorbox{palePurple}{0.726} & 0.319\\
OmniGen \cite{xiao2024omnigen}& 0.511  & 0.722 & \colorbox{paleRed}{\textbf{0.331}}\\
\textbf{UNO (Ours)} & \colorbox{paleRed}{\textbf{0.542}}  & \colorbox{paleRed}{\textbf{0.733}} & \colorbox{palePurple}{0.322}\\
\bottomrule
\end{tabular}
%}
\caption{\textbf{Quantitative results for multi-subject driven generation.} Our method achieves state-of-the-art performance among both tuning methods and tuning-free methods.}
\label{tab2:multi_ip}
\end{table}

\subsection{Qualitative Analyses}
We compare with various state-of-the-art methods to verify the effectiveness of our proposed UNO. We show the comparison of single-image condition generation results in \cref{fig4:single_ip}. In the first two rows, our UNO nearly perfectly keeps the subject detail (e.g., the numbers on the dial of the clock) in the reference image, while other methods struggle to maintain the details. In the following two rows, we demonstrate the editability. UNO can maintain subject similarity while editing attributes, specifically colors, whereas other methods either fail to maintain subject similarity or do not follow the text edit instructions. As a contrast, OminiControl~\cite{tan2024ominicontrol} has good retention ability but may encounter copy-paste risks, e.g., a red robot in the last row of \cref{fig4:single_ip}. We show the comparison results of multi-image condition generation in \cref{fig5:multi-ip}. Our method can keep all reference images while adhering to text responses, whereas other methods either fail to maintain subject consistency or miss the input text editing instructions.

\subsection{Quantitative Evaluations}
\noindent \textbf{Automatic scores. }\cref{tab1:single_ip} compares our UNO on DreamBench~\cite{ruiz2023dreambooth} against both tuning-based and tuning-free methods. UNO has a significant lead over previous methods with the highest DINO and CLIP-I scores of $0.760$ and $0.835$ respectively in zero-shot scenarios, and a leading CLIP-I score of $0.304$. We also compare our method in the multi-image condition scenario in \cref{tab2:multi_ip}. UNO achieves the highest DINO and CLIP-I scores and has competitive CLIP-T scores compared to existing leading methods. This shows that UNO can greatly improve subject similarity while adhering to text descriptions.

\begin{figure}[t]
\centering
\includegraphics[scale=0.32]{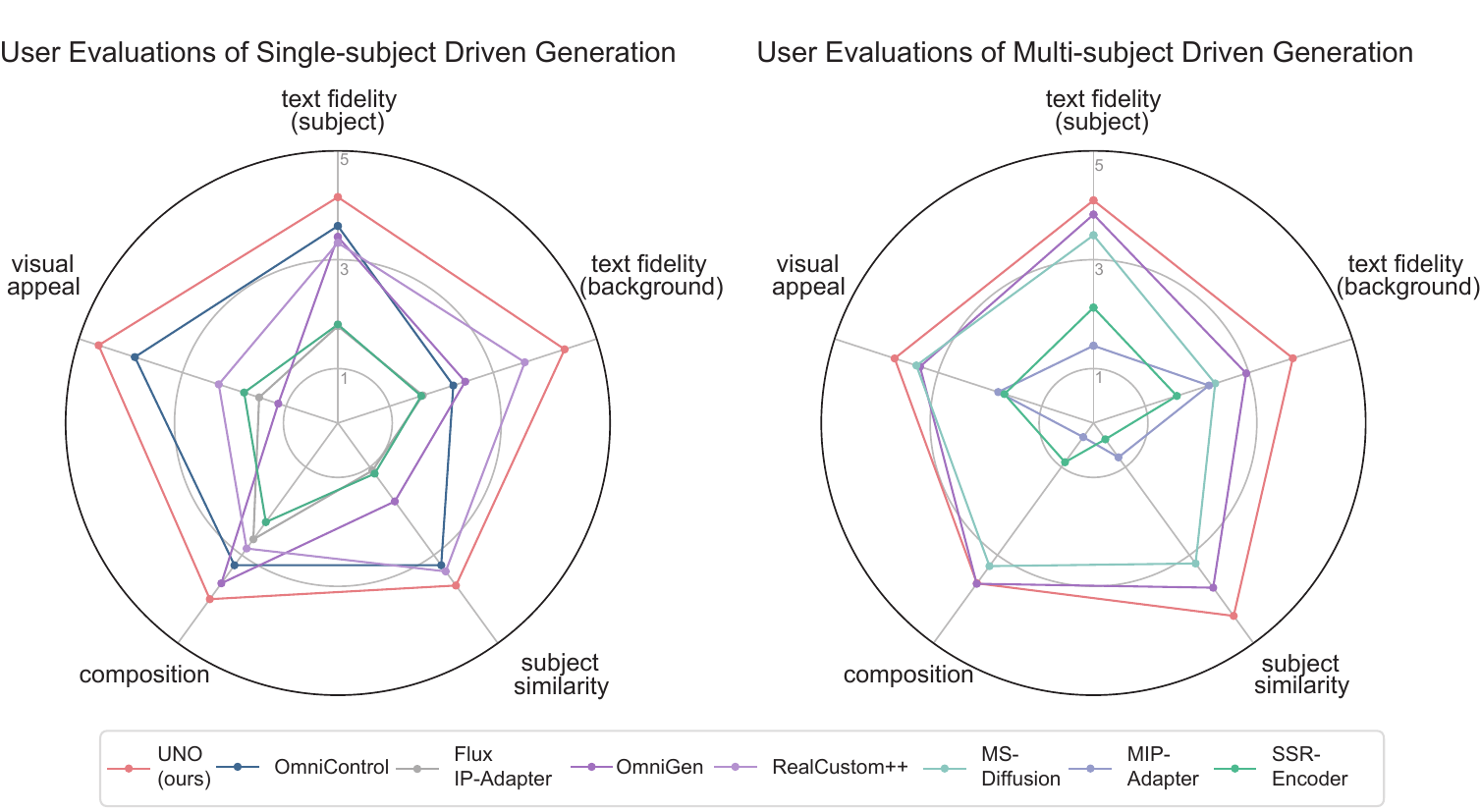}
    \caption{Radar charts of user evaluation of methods for single-subject driven and multi-subject driven generation on different dimensions}
    \label{fig:user_study}
\end{figure}

\noindent \textbf{User study. }We further conduct a user study via online questionnaires to showcase the superiority of UNO. For subjective assessment, $30$ evaluators including both domain experts and non-experts, assessed $300$ image combinations covering both single-subject and multi-subject driven generation tasks. For each case, evaluators rank the best results across five dimensions, including text fidelity at the subject level, text fidelity at the background level, subject similarity, composition quality, and visual appeal. As shown in \cref{fig:user_study}, the results reveal that our UNO not only excels in subject similarity and text fidelity but also achieves strong performance in other dimensions.

% \noindent \textbf{User study. }We further conduct a user study via online questionnaires to showcase the superiority of UNO. The results reveal that our UNO not only excels in subject similarity and text fidelity but also achieves strong performance in other dimensions. Further comprehensive comparisons can be found in our \textit{supplementary material}.

\subsection{Ablation Study}

\begin{table}[t]
\centering
\footnotesize
% \resizebox*{0.4\textwidth}{!}{
\begin{tabular}{lcccc}
\toprule
\textbf{Method} & \textbf{DINO} $\uparrow$ & \textbf{CLIP-I} $\uparrow$ & \textbf{CLIP-T} $\uparrow$ \\ \toprule
w/o generated $I_{\text{ref}}^2$ & 0.529 & 0.730 & 0.308\\
w/o cross-modal alignment & 0.511  & 0.721 & 0.322\\
w/o UnoPE & 0.386  & 0.674 & \textbf{0.323}\\\hline
\textbf{UNO (Ours)}  & \textbf{0.542} & \textbf{0.733} & 0.322\\
\bottomrule
\end{tabular}
%}
\caption{\textbf{Ablation Study of our proposed in-context data generation and in-context training method.} We report the results on the multi-subject driven generation benchmark.}
\label{tab3:ablation}
\end{table}

% in-context data generation
\begin{table}[t]
\centering
\footnotesize
% \resizebox*{0.4\textwidth}{!}{
\begin{tabular}{lcccc}
\toprule
\textbf{Method} & \textbf{DINO} $\uparrow$ & \textbf{CLIP-I} $\uparrow$ & \textbf{CLIP-T} $\uparrow$ \\ \toprule
w/ single subject pair-data& 0.730 & 0.821 & \textbf{0.309}\\
w/ cross-modal alignment & \textbf{0.760}  & \textbf{0.835} & 0.304\\
\bottomrule
\end{tabular}
% }
\caption{\textbf{Effect of progressive cross-modal alignment.} The model exhibits superior performance on DreamBench~\cite{ruiz2023dreambooth} after undergoing progressive cross-modal alignment, in contrast to being trained exclusively on single-subject pair-data, despite both models undergoing an identical number of training steps.}
\label{tab4:in-context}
\end{table}

\begin{figure}[t]
\centering
\includegraphics[scale=0.495]{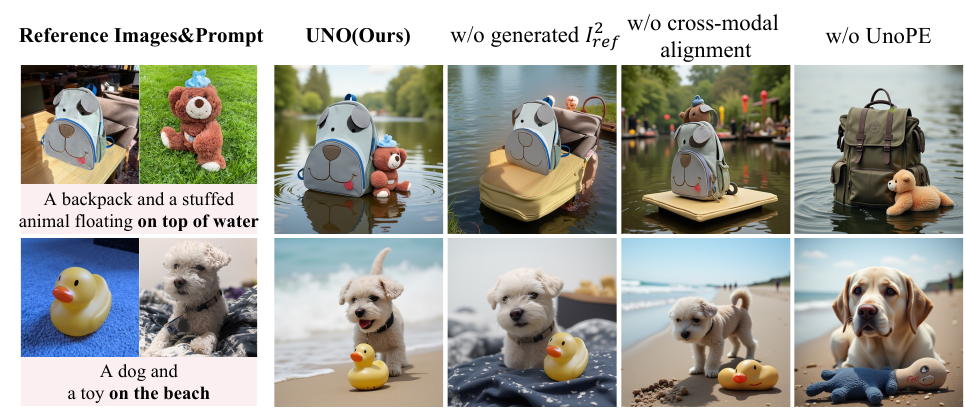}
    \caption{Ablation study of \textbf{UNO}. Zoom in for details.}
    \label{fig6:ab}
\end{figure}

\noindent \textbf{Effect of synthetic data curation framework. }\cref{tab3:ablation} shows the effects of different modules of UNO. When using augmented cropped part images from the target image instead of generated $I_{\text{ref}}^2$, we observe a significant decline in all metrics in \cref{tab3:ablation}. In \cref{fig6:ab}, the results tend to merely copy-paste the subjects and almost do not respond to the text prompt description.

\noindent \textbf{Effect of progressive cross-modal alignment. }As shown in \cref{tab3:ablation} and \cref{fig6:ab}, there is a significant drop in both DINO and CLIP-I scores, as well as in subject similarity, when the model is directly exposed to multiple reference image inputs without progressive cross-modal alignment. Furthermore, as shown in \cref{tab4:in-context}, progressive cross-modal alignment can increase the upper limit of the model in single-image condition scenarios.

% % dia pe single-ip+multi-ip, split
\begin{table}[t]
\centering
\resizebox*{0.4\textwidth}{!}{
\begin{tabular}{lcccc}
\toprule
\textbf{Method} & \textbf{DINO} $\uparrow$ & \textbf{CLIP-I} $\uparrow$ & \textbf{CLIP-T} $\uparrow$ \\ \toprule
w/o offset  & 0.470 & 0.722 & 0.308\\
w/ width-offset  & 0.717  & 0.813 & 0.304\\
w/ height-offset  & 0.678  & 0.797 & 0.308\\\hline
\textbf{UNO (Ours)} & \textbf{0.730}  & \textbf{0.821} & \textbf{0.309}\\
\bottomrule
\end{tabular}
}
\caption{Comparison with different forms of position index offsets. We report the results on DreamBench\cite{ruiz2023dreambooth}.}
\label{tab5:pe_single_ip}
\end{table}

\begin{table}[t]
\centering
\resizebox*{0.4\textwidth}{!}{
\begin{tabular}{lcccc}
\toprule
\textbf{Method} & \textbf{DINO} $\uparrow$ & \textbf{CLIP-I} $\uparrow$ & \textbf{CLIP-T} $\uparrow$ \\ \toprule
w/o offset & 0.386 & 0.674 & \textbf{0.323}\\
w/ width-offset  & 0.508  & 0.724 & 0.321\\
w/ height-offset  & 0.501  & 0.719 & 0.306\\\hline
\textbf{UNO (Ours)} & \textbf{0.542}  & \textbf{0.733} & 0.322\\
\bottomrule
\end{tabular}
}
\caption{Comparison with different forms of position index offsets. We report the results on the multi-subject driven generation benchmark.}
\label{tab6:pe_multi_ip}
\end{table}

\noindent \textbf{Effect of UnoPE. }As shown in \cref{tab3:ablation}, there is a significant drop in both DINO and CLIP-I scores when cloning the position index from the target image without using UnoPE. In \cref{fig6:ab}, the generated images can follow the text descriptions but hardly reference the input images. We further compared with different forms of position index offsets, as shown in \cref{tab5:pe_single_ip,tab6:pe_multi_ip}, and our method achieves the best results, which demonstrates the superiority of our proposed UnoPE.
\section{Conclusion}
\label{sec:Conclusion}
\label{sec:conclusion}
In this paper, we present UNO, a universal customization architecture that unlock the multi-condition contextual capabilities of diffusion transformer. This is achieved through progressive cross-modal alignment and universal rotary position embedding. The training of UNO consists of two steps. The first step uses single-image input to elicit the subject-to-image capabilities in diffusion transformers. The next step involves further training on multiple-subject data pairs. Our proposed universal rotary position embedding can also significantly improves subject similarity. Additionally, we present a progressive synthesis pipeline that evolves from single-subject to multi-subject in-context generation. This pipeline generates high-quality synthetic data, effectively reducing the copy-paste phenomenon. Extensive experiments show that UNO achieves high-quality similarity and controllability in both single-subject and multiple-subject customization.

{
    \small
    \bibliographystyle{ieeenat_fullname}
    \bibliography{main}
}

\clearpage
\setcounter{page}{12}

% 添加附录标题
\onecolumn
\section*{\centering\LARGE\bfseries Less-to-More Generalization: \\ Unlocking More Controllability by In-Context Generation}
\section*{\centering Supplementary Material}
% 自定义 section 和 subsection 的编号格式
\renewcommand{\thesection}{\Alph{section}}
\renewcommand{\thesubsection}{\Alph{section}.\arabic{subsection}}

\section{In-Context Data Generation Pipeline}\label{sup_sec:In-Context Data Generation Pipeline}
In this section, we give a detailed description of our in-context data generation pipeline. We first build a taxonomy tree in \cref{sup_subsec:Taxonomy Tree Generation} to obtain various subject instances and scenes. Then we generate subject-consistent image-pair data with the in-context ability of pretrained Text-to-Image (T2I) model and utilize Chain-of-Thought (CoT)~\cite{wei2022chain_cot} to filter the synthesized data in \cref{sup_subsec:Single-Subject In-Context Data Generation}. Finally, for multi-subject data, we train a Subject-to-Image (S2I) model to generate subject-consistent reference image instead of the cropped one to avoid the copy-paste issue in \cref{sup_subsec:Multi-Subject In-Context Data Generation}.

\subsection{Taxonomy Tree Generation}\label{sup_subsec:Taxonomy Tree Generation}
\begin{figure}[ht]
\centering
\includegraphics[width=0.95\linewidth]{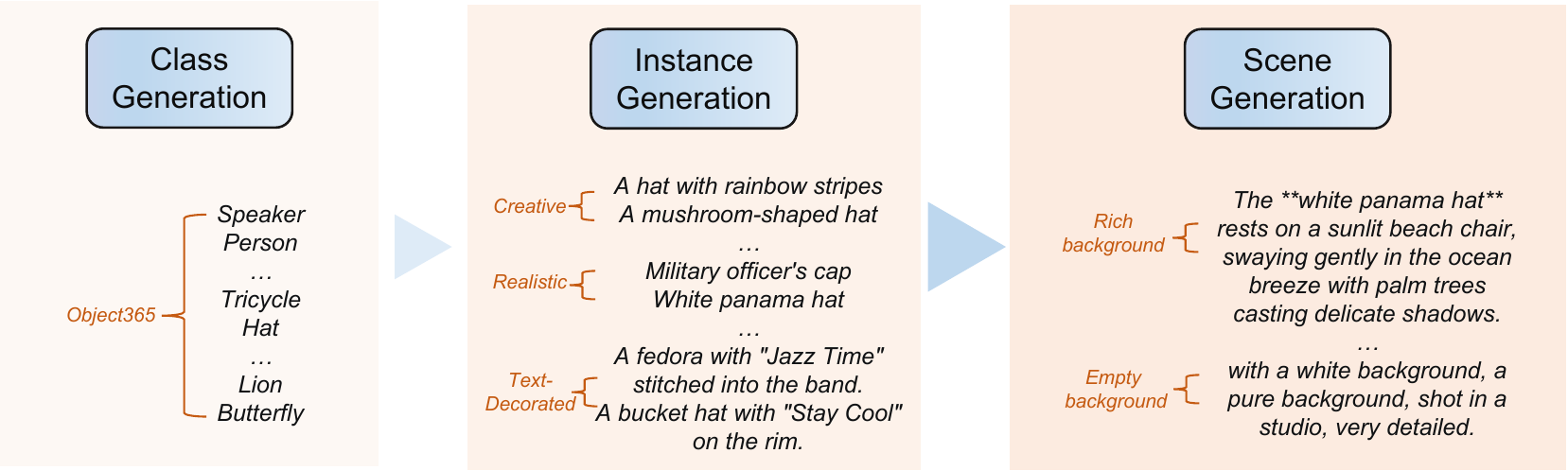}
    \caption{Illustration of the taxonomy tree.}
    \label{sup_fig1}
\end{figure}

To ensure the diversity of the generated dataset, we first construct a taxonomy tree that includes common categories of people and objects, as shown in \cref{sup_fig1}. Specifically, we use the 365 general classes from Object365~\cite{shao2019objects365} as the basis for our taxonomy tree. To obtain more diverse categories, we employ Large Language Model (LLM) to generate various subject instances and diverse scenes. The instructions in \cref{sup_fig:gen_ins} make LLM generate subject instances according to the given asset category in creative, realistic, and text-decorated ways. In addition, we instruct LLM to generate scene descriptions according to the given subject with system prompt in \cref{sup_fig5}. Following the steps above, we build a taxonomy tree and get plenty of diverse subjects and scene descriptions.

\subsection{Single-Subject In-Context Data Generation}\label{sup_subsec:Single-Subject In-Context Data Generation}
In-context ability of T2I model has been proved in OminiControl~\cite{tan2024ominicontrol} and we also utilize it to generate subject-consistent image-pair data as the basic part of our final in-context synthesized data in \cref{sup_subsubsec:Subject-Consistent Image-Pair Generation}. Moreover, we filter out low quality data with bad subject consistency according to the similarity from DINOv2~\cite{oquab2023dinov2} and Vision-Language Model (VLM) in \cref{sup_subsubsec:Subject-Consistent Image-Pair Filter}.

\begin{figure}[ht]
\centering
\includegraphics[width=0.5\linewidth]{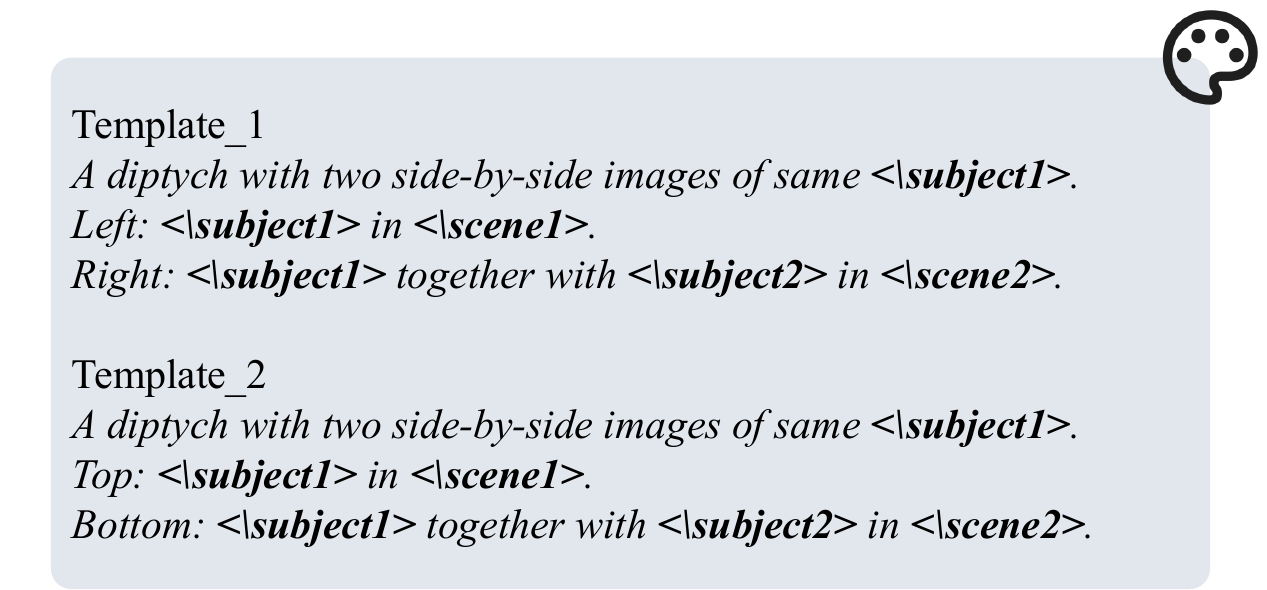}
    \caption{Diptych text template for generating subject-consistent image-pair with FLUX.1\cite{blackforestlabs_flux}.}
    \label{sup_fig6}
\end{figure}

\subsubsection{Subject-Consistent Image-Pair Generation}\label{sup_subsubsec:Subject-Consistent Image-Pair Generation}
Combining the constructed taxonomy tree with the predifined diptych text template in \cref{sup_fig6}, we utilize the inherent in-context ability of FLUX.1~\cite{blackforestlabs_flux}, one of the state-of-the-art T2I model, to generate subject-consistent image-pair. Since FLUX.1~\cite{blackforestlabs_flux} has multi-resolution generation ability, we directly produce three different high-resolution (\ie, 1024$\times$1024, 1024$\times$768, 768$\times$1024) image-pairs, with great balance of quality and efficiency.

\subsubsection{Subject-Consistent Image-Pair Filter}\label{sup_subsubsec:Subject-Consistent Image-Pair Filter}
Though FLUX.1~\cite{blackforestlabs_flux} shows great in-context generation ability, synthesized image-pairs suffer several issues, especially subject inconsistency and missing subjects. We highlight that the high quality of synthesized data can notably accelerate the convergence and improve the subject consistency. To efficiently filter synthesized data, we first split the diptych image-pair into reference image $I_{\text{ref}}^1$ and target image $I_{\text{tgt}}$ with Hough Transform. According to the template in \cref{sup_fig6}, both $I_{\text{ref}}^1$ and $I_{\text{tgt}}$ contain the same \textit{subject1} while $I_{\text{tgt}}$ has another \textit{subject2}. To ensure $I_{\text{ref}}^1$ and $I_{\text{tgt}}$ have consistent \textit{subject1}, we then calculate cosine similarity with DINOv2~\cite{oquab2023dinov2} and set a threshold to filter out image-pairs with significantly low consistency.

However, since the reference image $I_{\text{ref}}^1$ and the target image $I_{\text{tgt}}$ have different scene settings, the \textit{subject1} in the image-pairs may not be spatially aligned, resulting in incorrect cosine similarity with feature from DINOv2~\cite{oquab2023dinov2}. We further employ VLM to provide a fine-grained score list evaluating various aspects adaptively, \ie appearance, details, and attributes. We only keep the data with highest VLM score, which indicating the highest quality and subject consistency in the synthesized data. Specifically, inspired by \cite{cai2024diffusion_selfdistill}, we utilize CoT~\cite{wei2022chain_cot} for better discrimination of the \textit{subject1} in $I_{\text{ref}}^1$ and $I_{\text{tgt}}$, as shown in \cref{sup_fig:cot}. To demonstrate the effectiveness of the CoT filter, we sample data from different VLM score intervals in \cref{sup_fig:cot_showcase}. Image-pairs with low score suffer severe subject inconsistency while those with highest score (\ie score is 4) show highly consistent subject in the reference image and the target image. We also count the amount of data in each score interval as shown in \cref{sup_fig:cot_score}, indicating that around 35.43\% data would be remained with the VLM CoT filter. Also, there are seldom of data with extremely low VLM score after DINOv2 filter, showing its effectiveness.

\begin{figure*}[h!]
    \centering
    \begin{subfigure}[b]{0.3\linewidth}
        \centering
        \includegraphics[width=\linewidth]{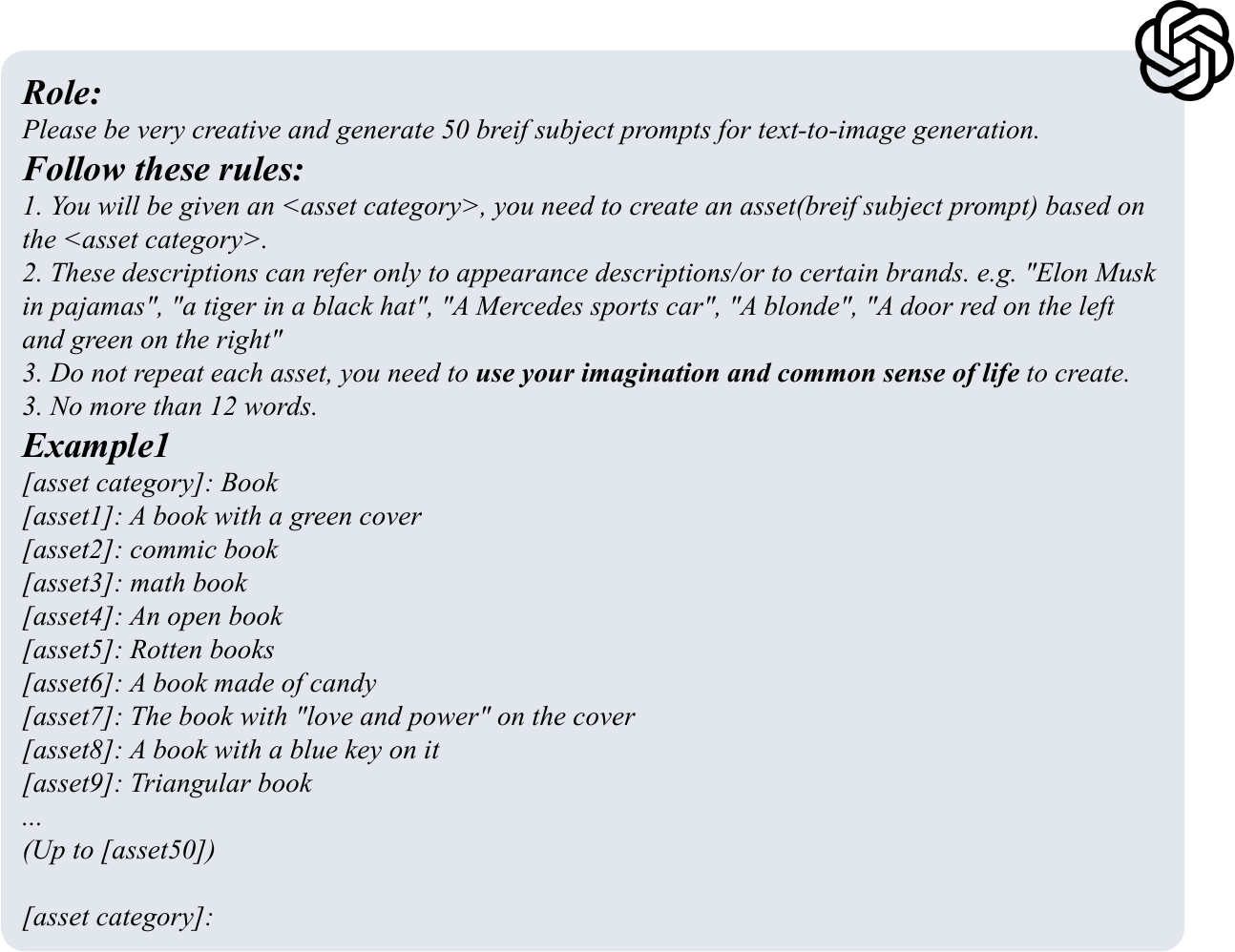}
        \caption{System prompt of LLM used to generate subject instances in creative type.}
        \label{sup_fig2}
    \end{subfigure}
    \hfill
    \begin{subfigure}[b]{0.3\linewidth}
        \centering
        \includegraphics[width=\linewidth]{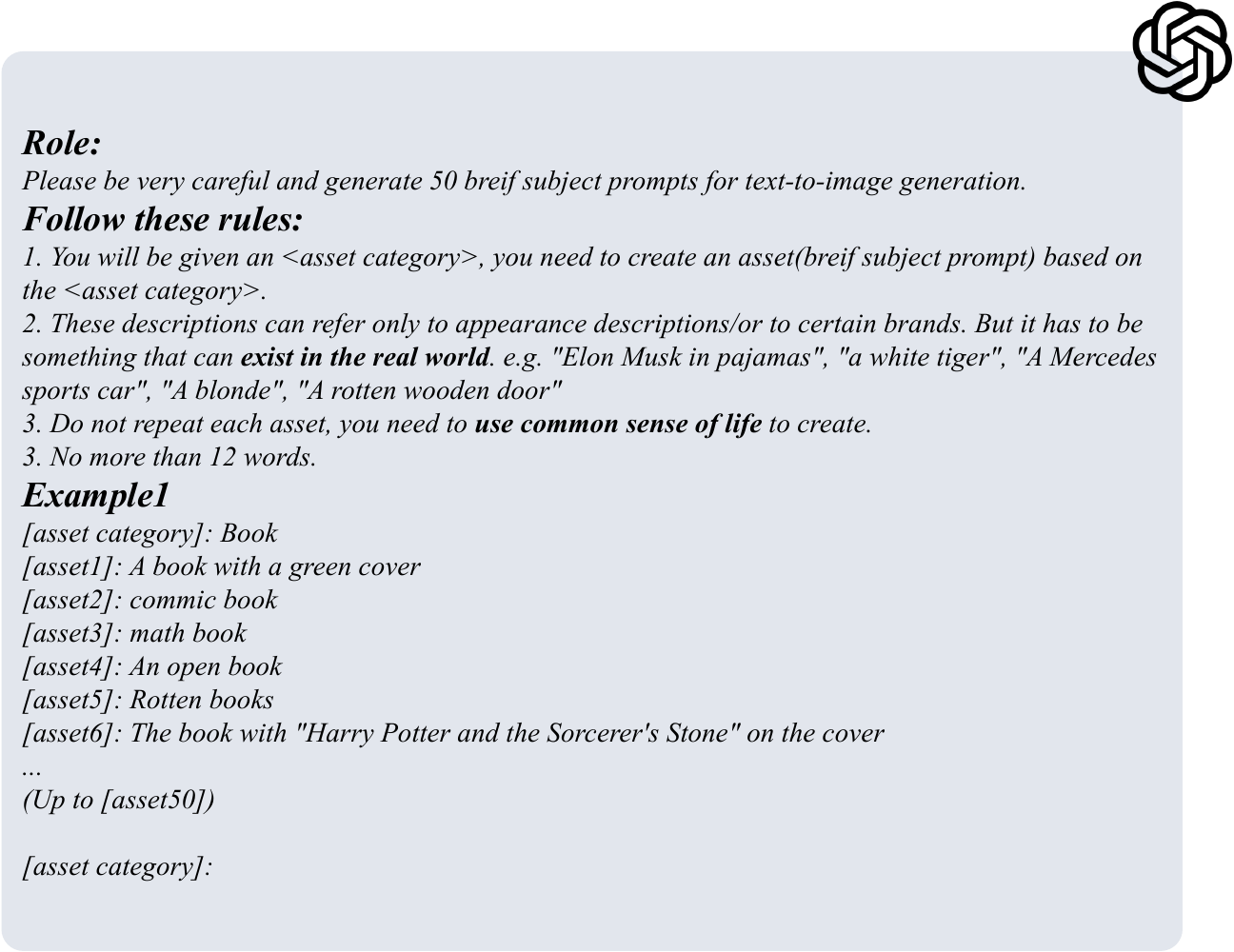}
        \caption{System prompt of LLM used to generate subject instances in realistic type.}
        \label{sup_fig3}
    \end{subfigure}
    \hfill
    \begin{subfigure}[b]{0.3\linewidth}
        \centering
        \includegraphics[width=\linewidth]{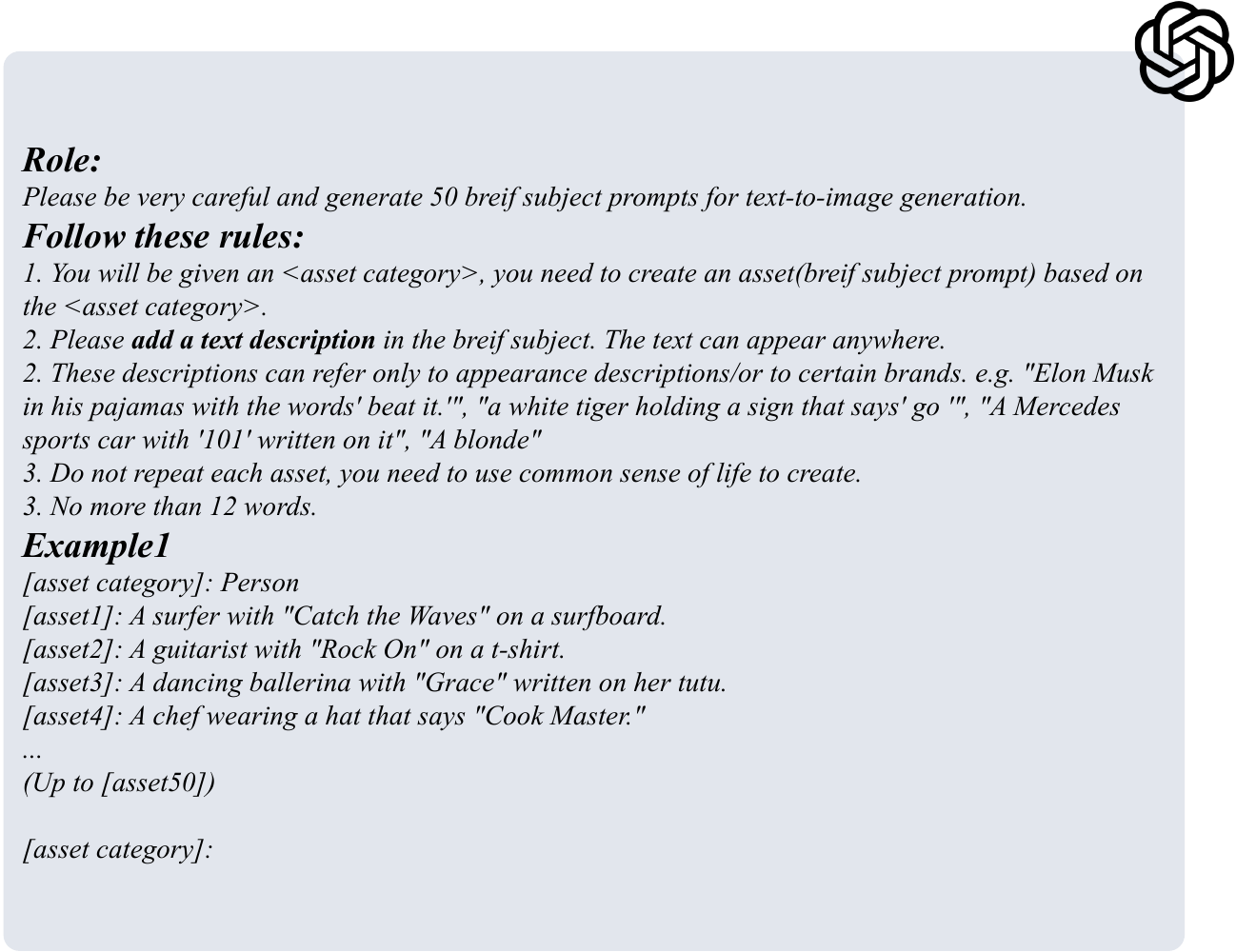}
        \caption{System prompt of LLM used to generate subject instances in text-decorated type.}
        \label{sup_fig4}
    \end{subfigure}
    \caption{System prompt of LLM used to generate subject instances.}
    \label{sup_fig:gen_ins}
\end{figure*}

\begin{figure*}[h!]
\centering
\includegraphics[width=1\linewidth]{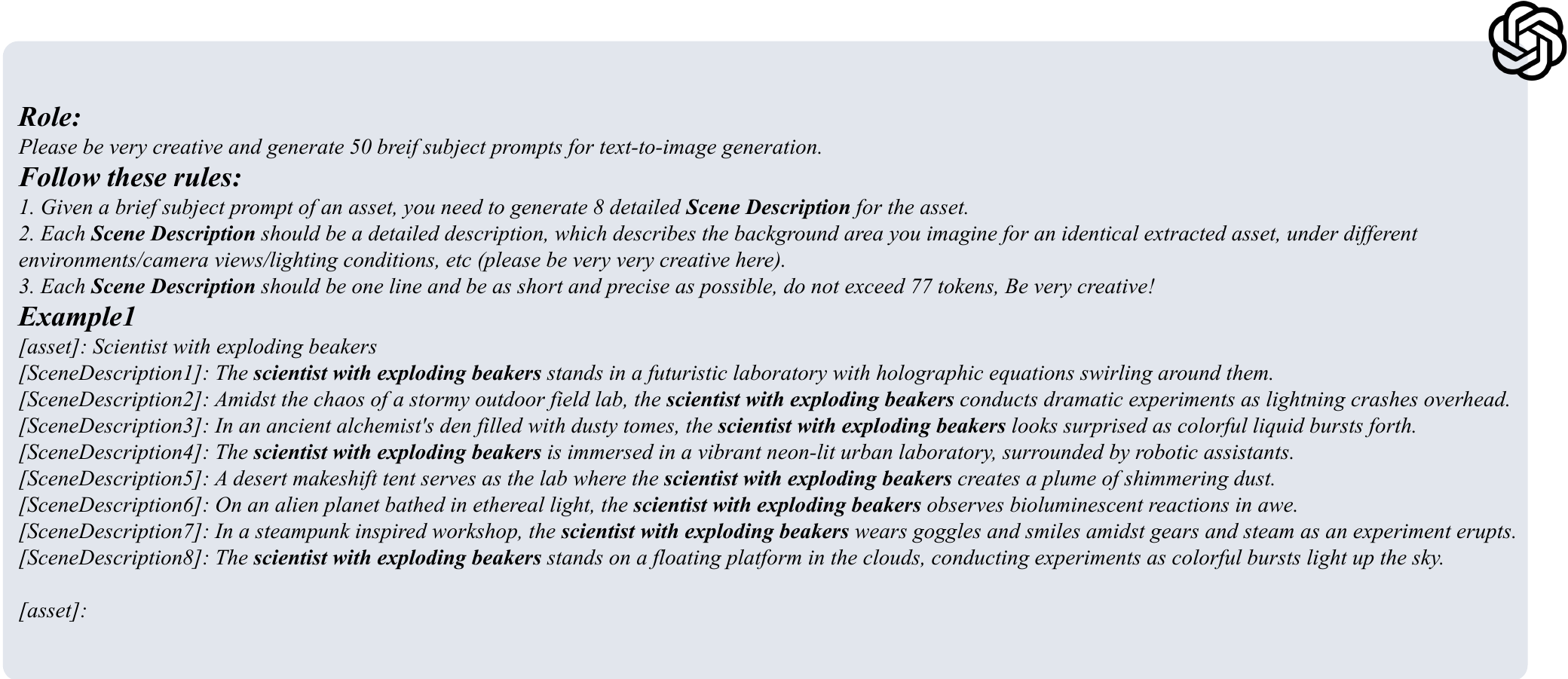}
    \caption{System prompt of LLM used to generate scene descriptions.}
    \label{sup_fig5}
\end{figure*}

\subsection{Multi-Subject In-Context Data Generation}\label{sup_subsec:Multi-Subject In-Context Data Generation}
Following the above pipeline, we construct single-subject in-context data containing subject-consistent image-pairs $(I_{\text{ref}}^1, I_{\text{tgt}})$. Both the reference image and the target image have \textit{subject1} while only the target image contain \textit{subject2}. Since $I_{\text{tgt}}$ has multi-subject, the simplest way to build multi-subject in-context data is utilizing open-vocabulary detector (OVD) to identify and crop the \textit{subject2} in $I_{\text{tgt}}$ as the second reference image $\bar I_{\text{ref}}^2$. However, we find that the cropped $\bar I_{\text{ref}}^2$ would make severe copy-paste issue. To alleviate the issue, a S2I model is trained with the single-subject in-context data and then used to generate new reference image $I_{\text{ref}}^2$, which has the consistent subject with the cropped $\bar I_{\text{ref}}^2$ but different scenes. Thus we have high quality multi-subject in-context data with subject-consistent image-pairs $(I_{\text{ref}}^1, I_{\text{ref}}^2, I_{\text{tgt}})$ after very similar filter pipeline for the synthesized $I_{\text{ref}}^2$. There are some case randomly sample from our final data in \cref{sup_fig:data_showcase}. Interestingly, we find that a small part of image-pairs have more than 2 reference images, due to the randomness of T2I generation and OVD, empowering generalization for more-subject generation.

\begin{figure*}[h!]
    \centering
    \begin{subfigure}[b]{0.45\linewidth}
        \centering
        \includegraphics[width=\linewidth]{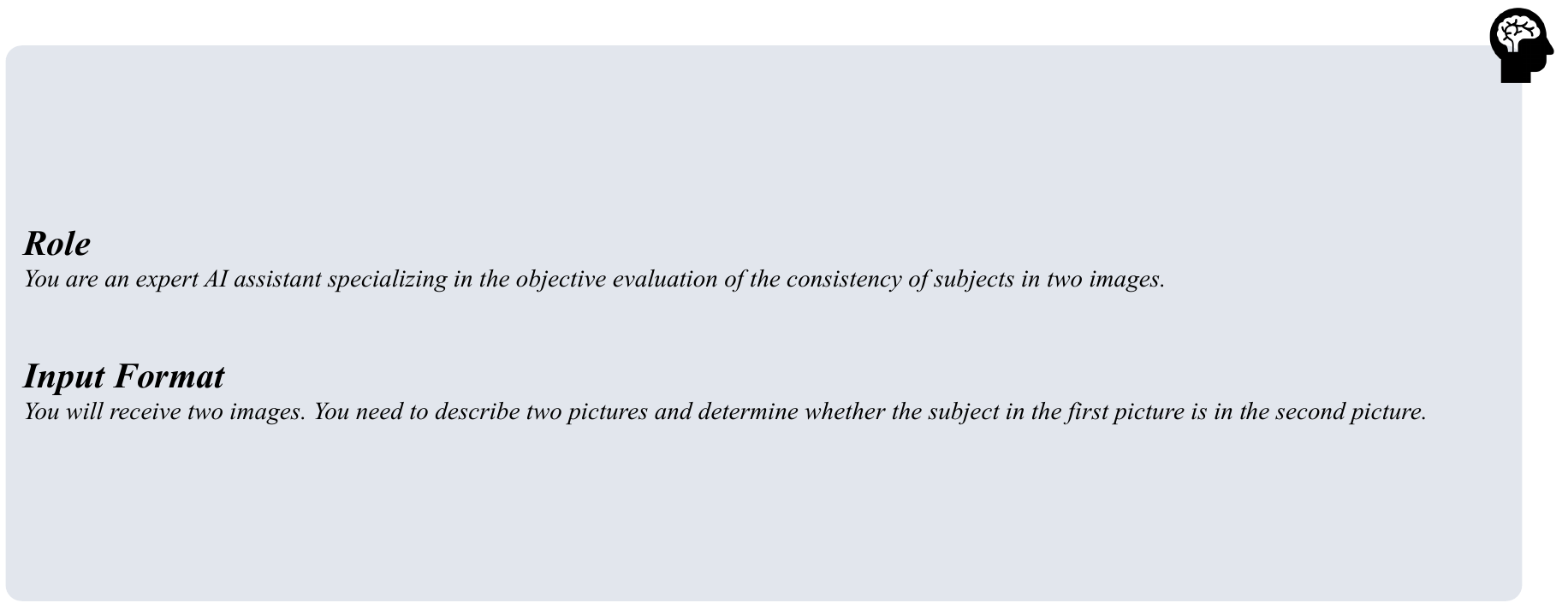}
        \caption{System prompt of the filter VLM.}
        \label{sup_fig:cot_sys}
    \end{subfigure}
    \hfill
    \begin{subfigure}[b]{0.45\linewidth}
        \centering
        \includegraphics[width=\linewidth]{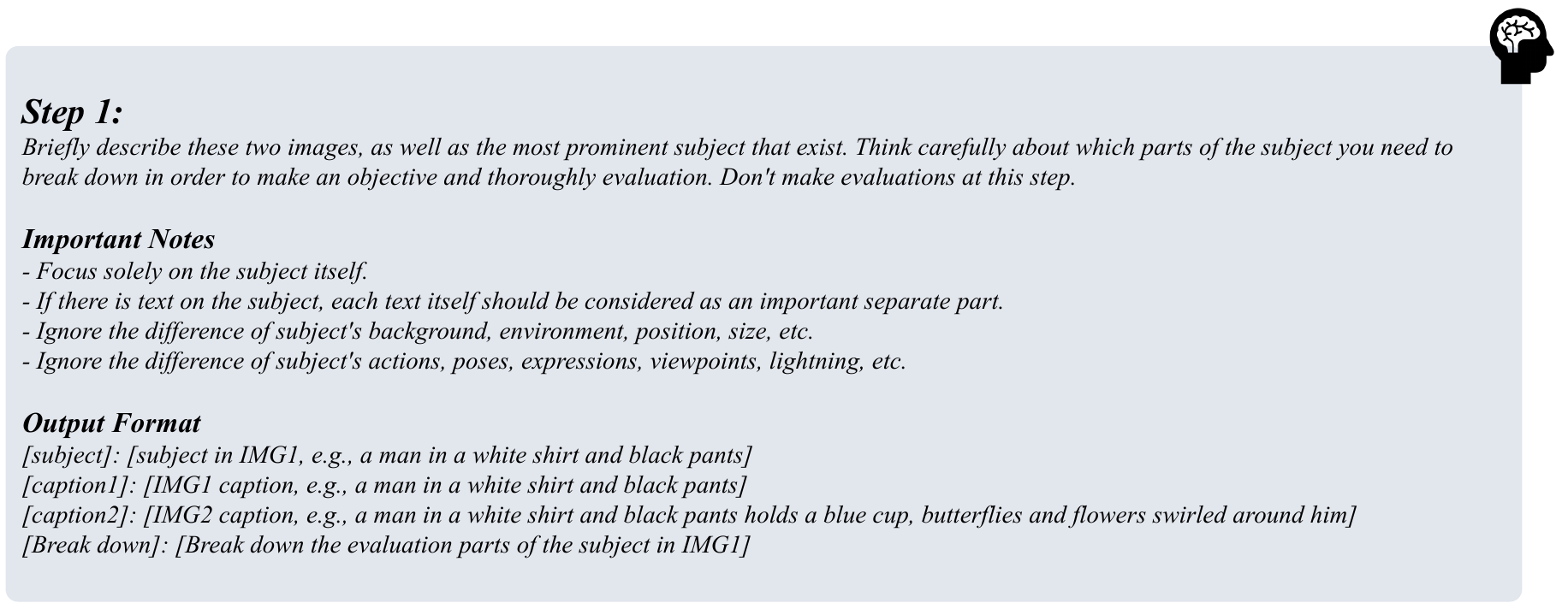}
        \caption{Prompt for the first round CoT of the filter VLM.}
        \label{sup_fig:cot_1}
    \end{subfigure}
    \hfill
    \begin{subfigure}[b]{0.45\linewidth}
        \centering
        \includegraphics[width=\linewidth]{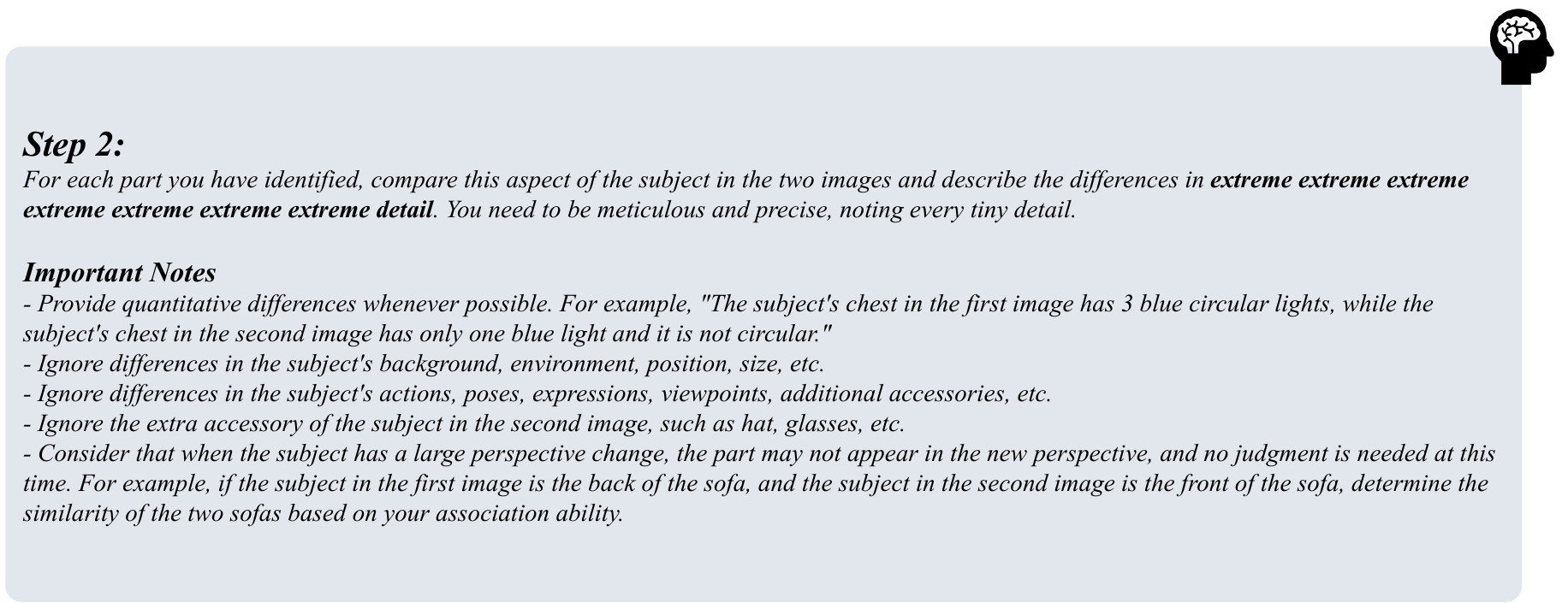}
        \caption{Prompt for the second round CoT of the filter VLM.}
        \label{sup_fig:cot_2}
    \end{subfigure}
    \hfill
    \begin{subfigure}[b]{0.45\linewidth}
        \centering
        \includegraphics[width=\linewidth]{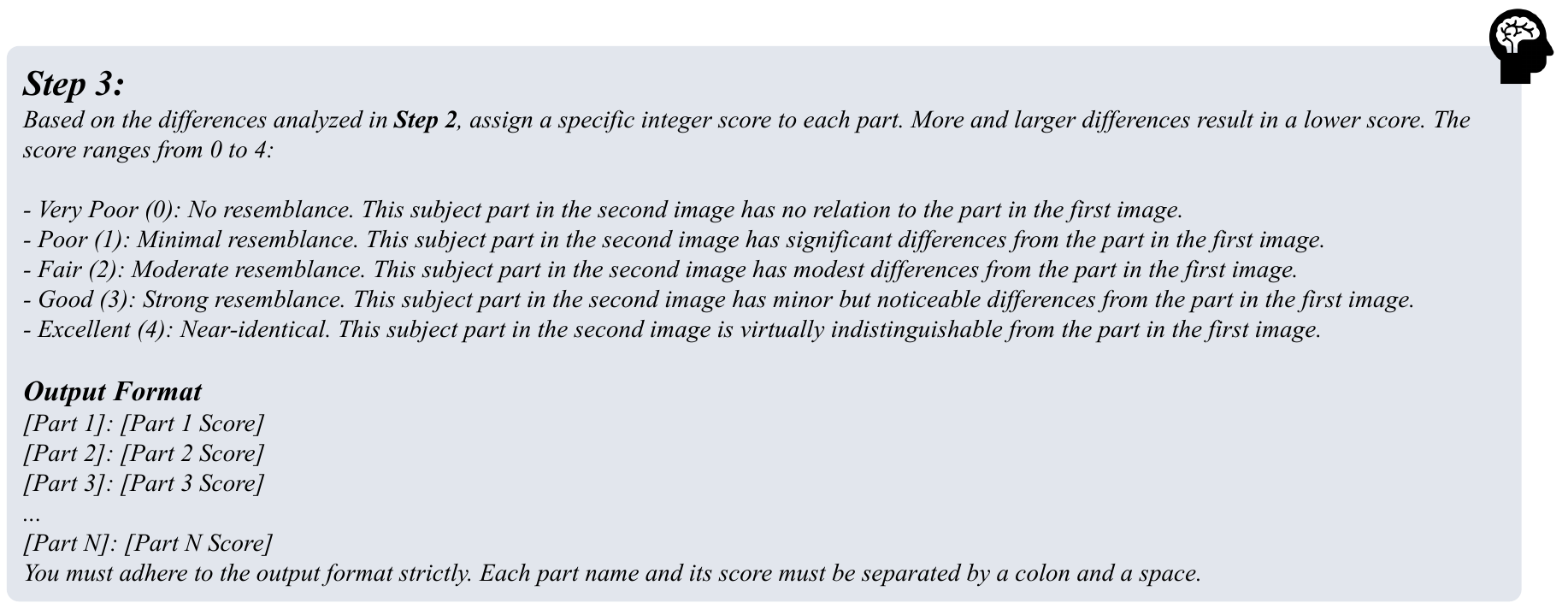}
        \caption{Prompt for the third round CoT of the filter VLM.}
        \label{sup_fig:cot_3}
    \end{subfigure}
    \caption{CoT prompt of the filter VLM.}
    \label{sup_fig:cot}
\end{figure*}

\begin{figure*}[h!]
\centering
\includegraphics[width=1\linewidth]{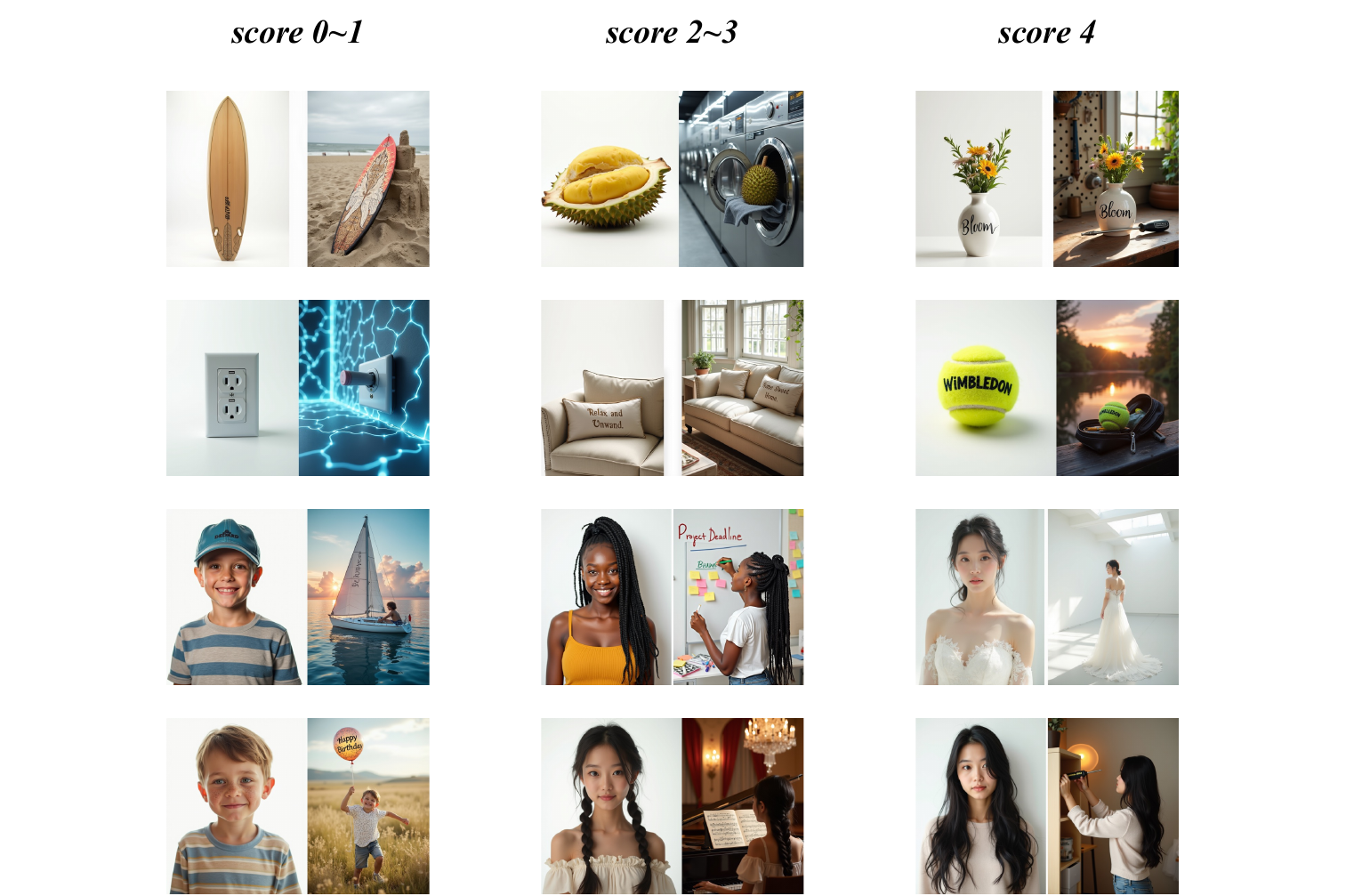}
    \caption{Sampled data from different VLM score intervals.}
    \label{sup_fig:cot_showcase}
\end{figure*}

\begin{figure*}[h!]
\centering
\includegraphics[width=1\linewidth]{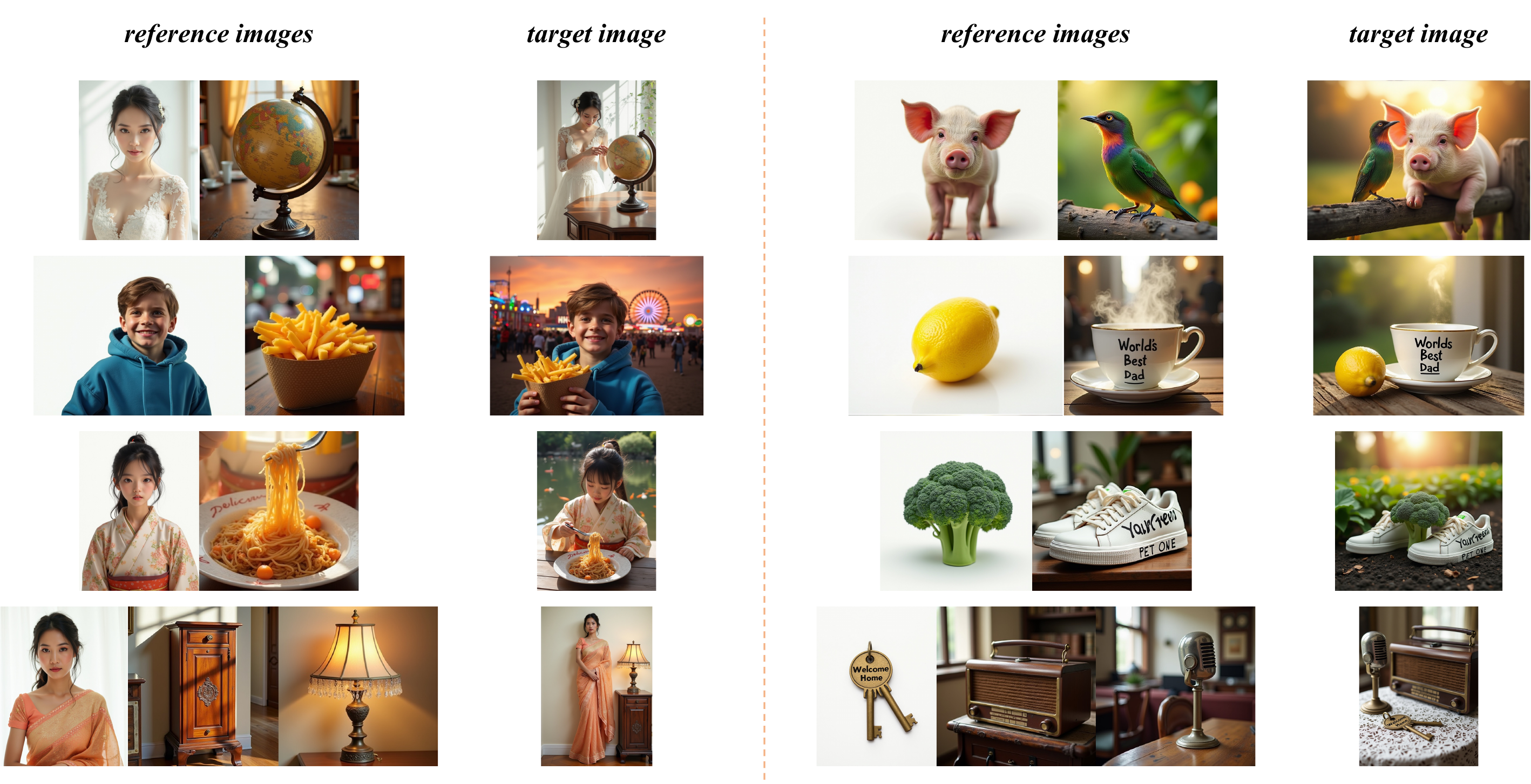}
    \caption{Sampled data from our final multi-subject in-context data.}
    \label{sup_fig:data_showcase}
\end{figure*}

\begin{figure*}[h!]
\centering
\includegraphics[width=0.35\linewidth]{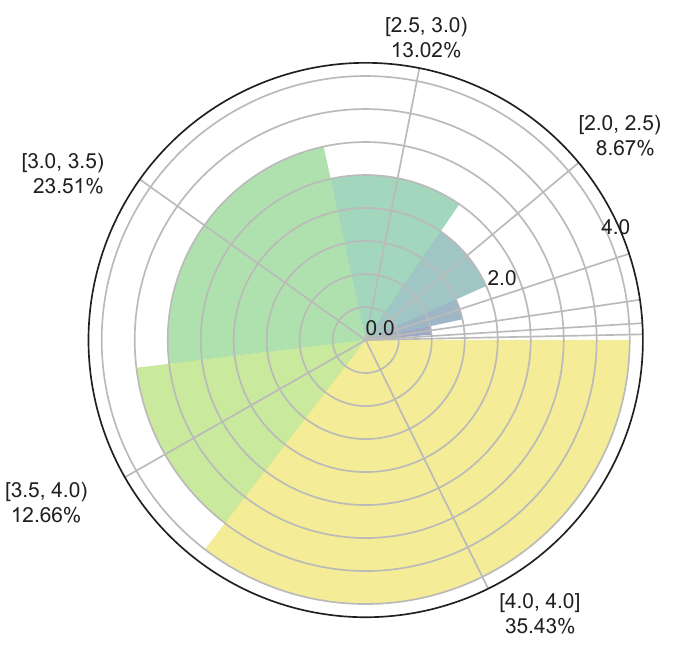}
    \caption{Amount of data in each VLM score interval.}
    \label{sup_fig:cot_score}
\end{figure*}

% \section{More Quantitative Results} % lora rank
\section{Analysis on LoRA Rank}
\begin{figure*}[h!]
\centering
\includegraphics[width=0.6\linewidth]{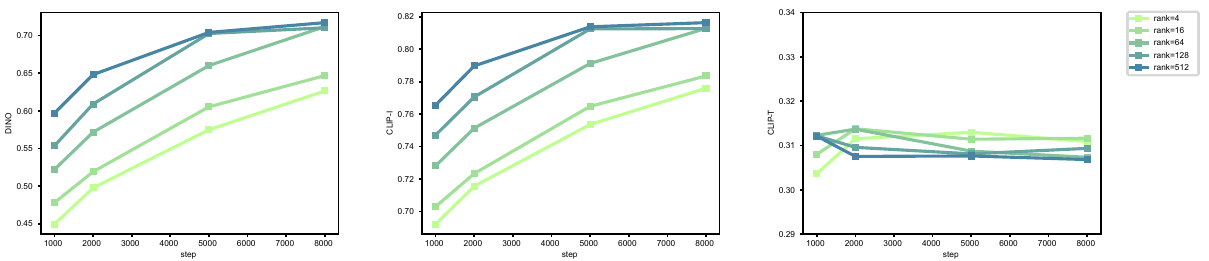}
    \caption{Analysis of model performance under different LoRA ranks.}
    \label{score_step_rank}
\end{figure*}

In this section, we further conduct an ablation study on the LoRA rank. Since training parameters are strongly related to the final performance, we scale the rank from 4 to 512. As shown in the \cref{score_step_rank}, increasing the rank gradually brings sustained gains, but when the rank reaches 128, the performance improvement slows down. Finally, considering both performance and resource consumption, we set UNO to a rank of 512.

\section{More Qualitative Results} % showcase, more-subject generation
\subsection{Qualitative Results on Multi-Subject Driven Generation}
\begin{figure*}[h!]
\centering
\includegraphics[width=0.95\linewidth]{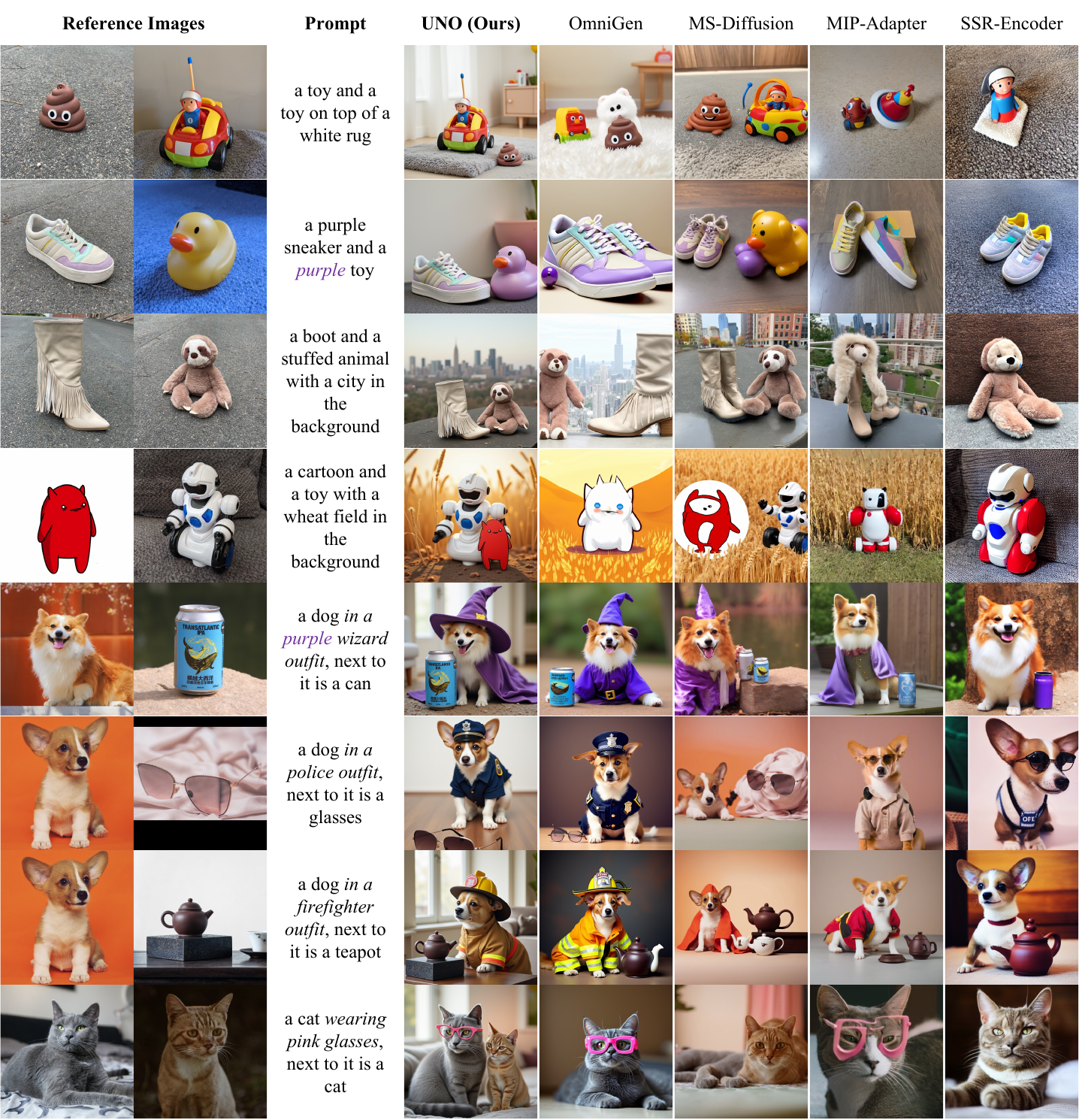}
    \caption{More comparison with different methods on multi-subject driven generation. We \textit{italicize} the subject-related editing part of the prompts.}
    \label{sup_fig:multi_ip_showcase}
\end{figure*}

We show a more qualitative comparison of multi-subject driven generation in \cref{sup_fig:multi_ip_showcase}. It is clear that UNO generate images with best multi-subject consistency, following edit instructions to the subject and background.

\subsection{Application Scenarios}
We evaluated our UNO model across diverse multi-image conditional scenarios, such as identity preservation, virtual try-on, and stylized generation. We found that UNO demonstrated exceptional generalization capabilities, even with minimal exposure to such data during training.

\begin{figure*}[h!]
\centering
\includegraphics[width=0.95\linewidth]{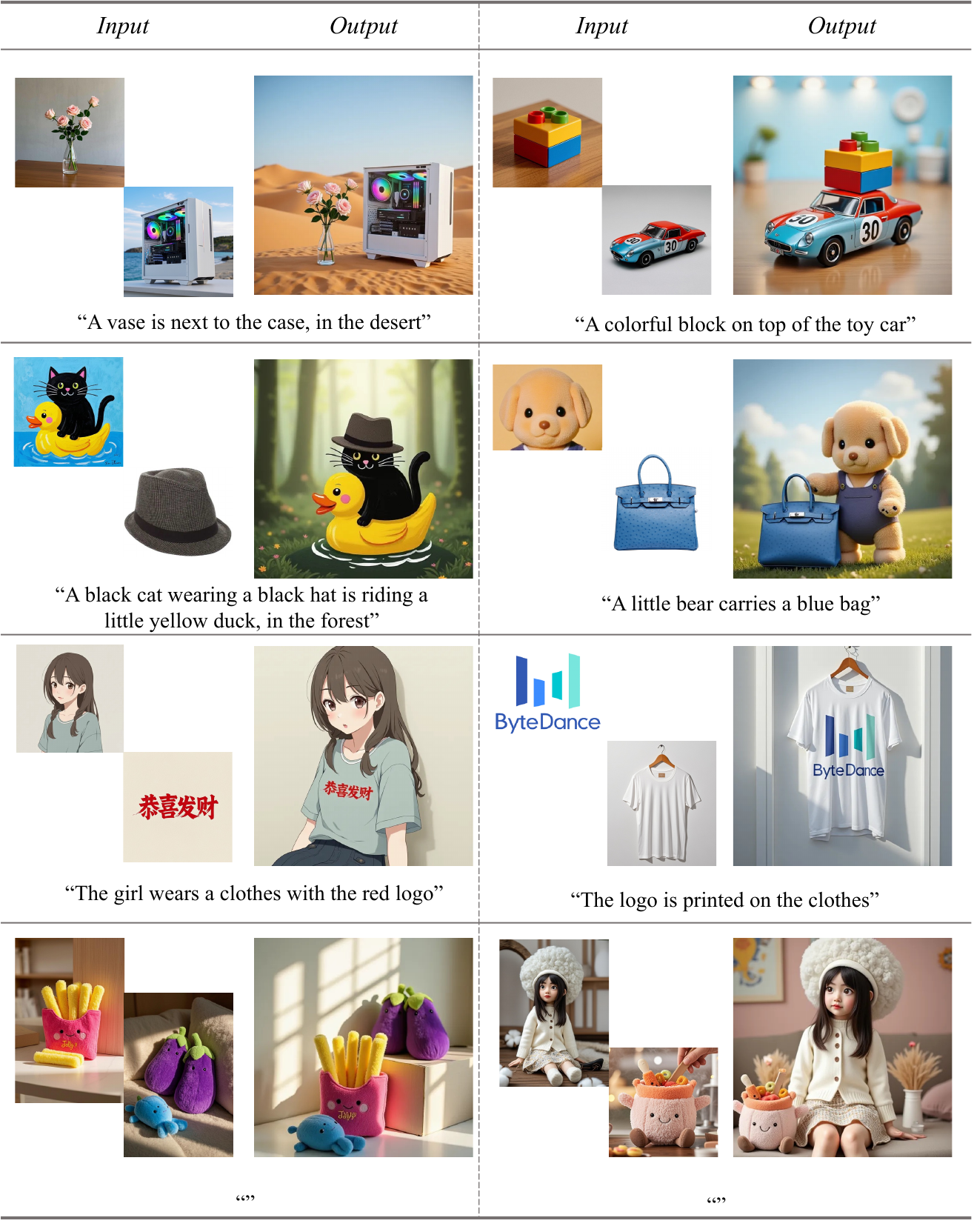}
    \caption{More multi-subject generation results from our UNO model.}
    \label{sup_fig:application1}
\end{figure*}

\begin{figure*}[h!]
\centering
\includegraphics[width=0.95\linewidth]{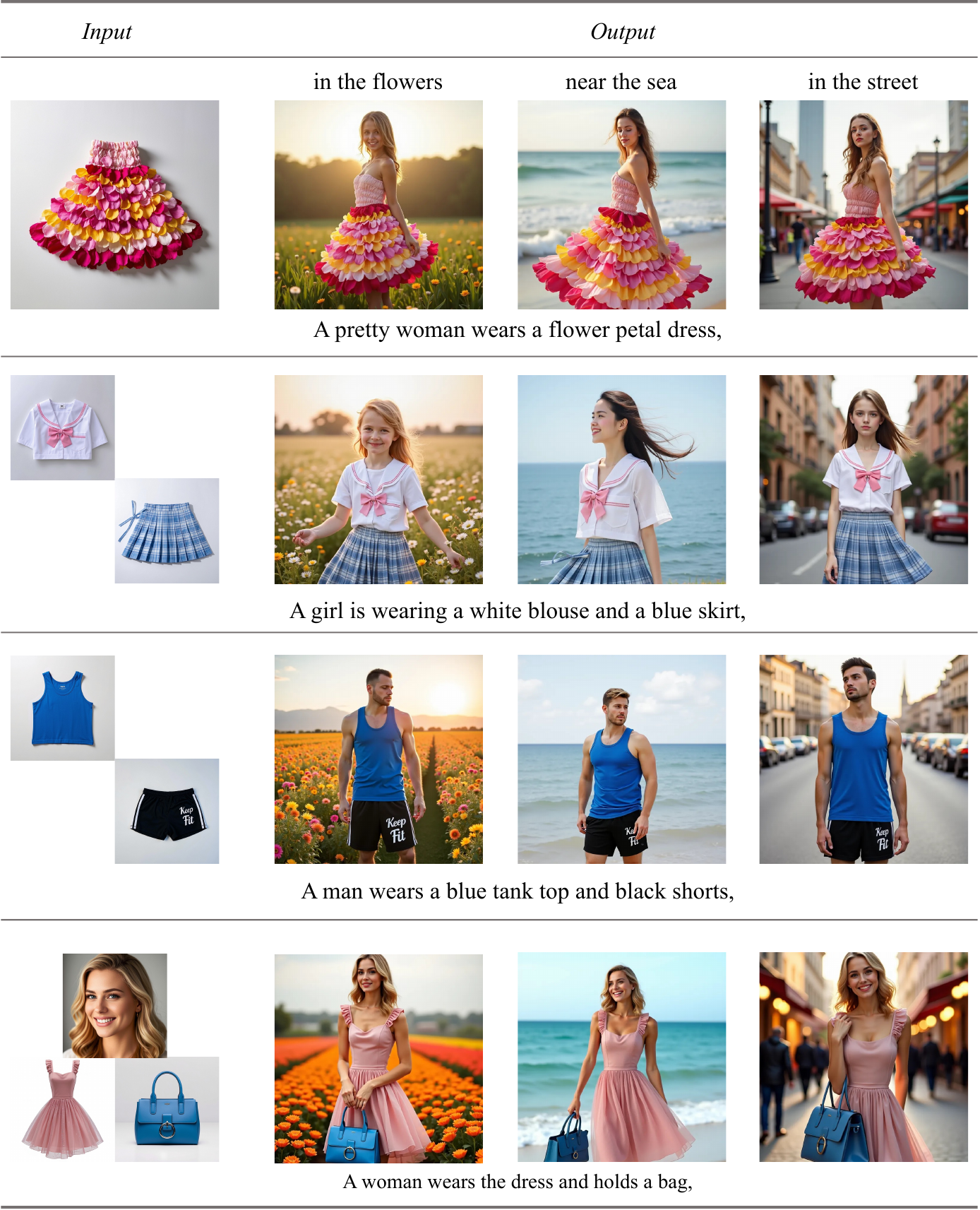}
    \caption{More virtual try-on results from our UNO model.}
    \label{sup_fig:application2}
\end{figure*}

\begin{figure*}[h!]
\centering
\includegraphics[width=0.95\linewidth]{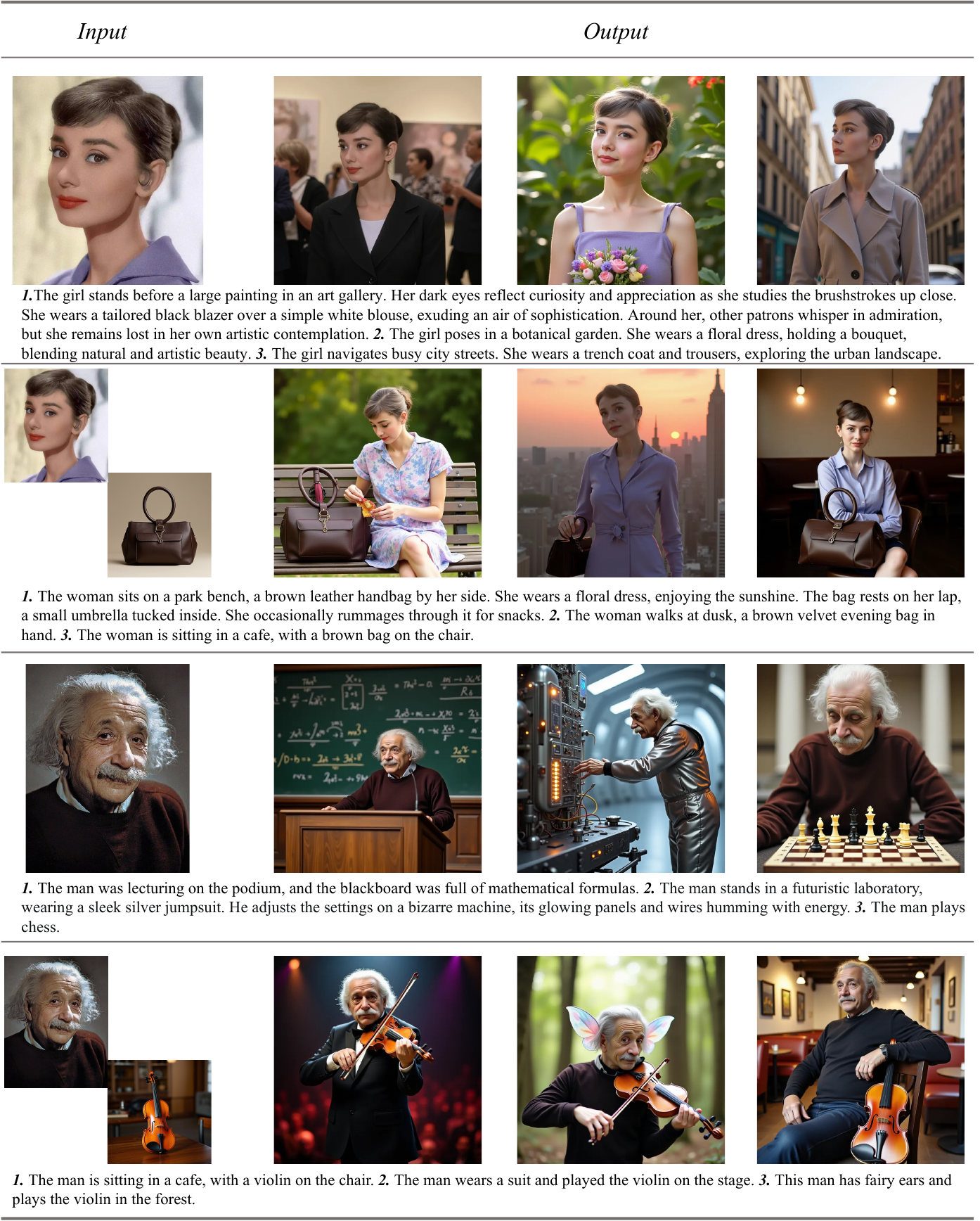}
    \caption{More identity preservation results from our UNO model.}
    \label{sup_fig:application3}
\end{figure*}

\begin{figure*}[h!]
\centering
\includegraphics[width=0.95\linewidth]{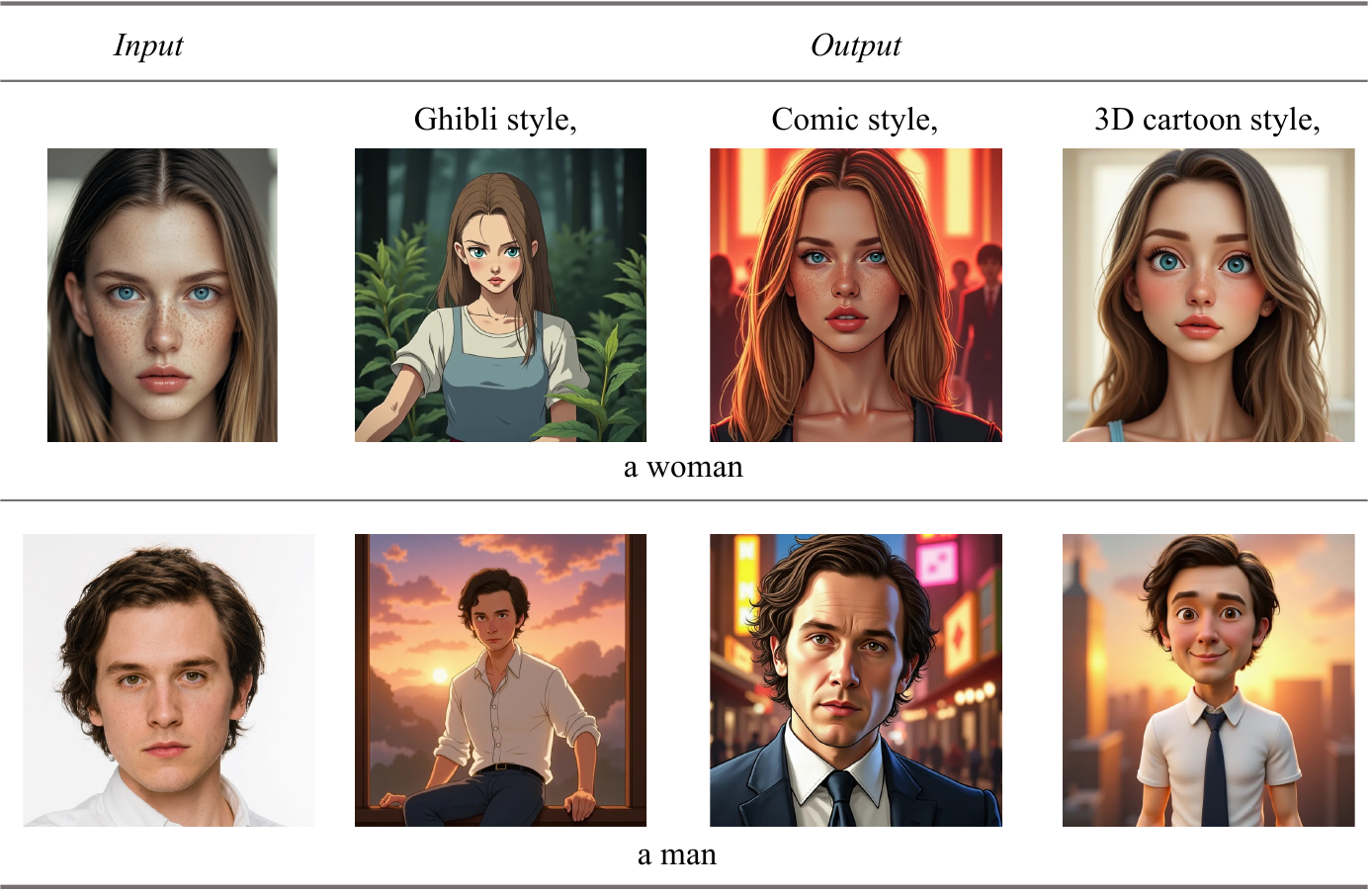}
    \caption{More stylized generation results from our UNO model.}
    \label{sup_fig:application4}
\end{figure*}

\textbf{\textit{Multi-subject Driven Generation: }}we have showcased additional results from our UNO model in \cref{sup_fig:application1}. Beyond effectively handling multi-subject scenarios, UNO excels in complex applications like logo design and the integration of virtual and real elements, demonstrating its strong generalization capabilities. \textbf{\textit{Virtual Try-on: }}as shown in \cref{sup_fig:application2}, UNO performs exceptionally well in virtual try-on scenarios, despite the absence of specialized training on such datasets. This demonstrates that UNO has learned to understand relationships between objects rather than simply performing copy-paste operations. It also suggests that UNO could provide novel optimization strategies for virtual try-on applications, a promising direction we leave to further exploration. \textbf{\textit{Identity Preservation: }}another notable observation is that UNO performs well in both pure ID scenarios and ID-subject combinations in \cref{sup_fig:application3}. This flexibility reduces reliance on additional ID plugins, fostering open-source community development. We attribute this capability to our systematic training data construction. As mentioned in \cref{sup_subsec:Taxonomy Tree Generation}, our taxonomy tree covers extensive human-object combinations, enabling this versatile performance. \textbf{\textit{Stylized Generation: }}as depicted in \cref{sup_fig:application4}, UNO has inherited stylization ability inherited from the original DiT model, despite the lack of specific paired data in our training set. This stems from our training approach, which smoothly transitions from T2I to S2I, allowing the model to evolve multi-condition control while maintaining strong semantic alignment.
% \subsection{User Study}

% \begin{figure}[ht]
% \centering
% \includegraphics[scale=0.32]{figs/user_study.pdf}
% % \includegraphics[scale=0.2]{figs/fig6_ablation.pdf}
%     \caption{Radar charts of user evaluation of methods for single-subject driven and multi-subject driven generation on different dimensions}
%     \label{fig:user_study}
% \end{figure}

% We further conduct a user study via online questionnaires to showcase the superiority of UNO. For subjective assessment, $30$ evaluators including both domain experts and non-experts, assessed $300$ image combinations covering both single-subject and multi-subject driven generation tasks. For each case, evaluators rank the best results across five dimensions, including text fidelity at the subject level, text fidelity at the background level, subject similarity, composition quality, and visual appeal. As shown in \cref{fig:user_study}, the results reveal that our UNO not only excels in subject similarity and text fidelity but also achieves strong performance in other dimensions.

% applications

\begin{table*}[t]
    \centering
    \renewcommand{\arraystretch}{0.9} % 调整行间距
    \begin{tabular}{lc}
        \toprule
        \textbf{Scenarios} & \textbf{Prompt} \\
        \midrule
        One2One & ``A clock on the beach is under a red sun umbrella" \\
                & ``A doll holds a `UNO' sign under the rainbow on the grass" \\
        \midrule
        Two2One & ``The figurine is in the crystal ball" \\
              & ``The boy and girl are walking in the street" \\
        \midrule
        Many2One & ``A penguin doll, a car and a pillow are scattered on the bed" \\
              & ``A boy in a red hat wear a sunglasses" \\    
        \midrule
        Stylized Generation & ``Ghibli style, a woman" \\
              & ``Ghibli style, a man" \\    
        \midrule
        Virtual Try-on & ``A man wears the black hoodie and pants" \\
              & ``A girl wears the blue dress in the snow" \\    
        \midrule
        Product Design & ``The logo and words `Let us unlock!' are printed on the clothes" \\
              & ``The logo is printed on the cup" \\    
        \midrule
        Identity-preservation & ``The figurine is in the crystal ball" \\
              & ``A penguin doll, a car and a pillow are scattered on the bed" \\    
        \midrule
        Story Generation & ``A boy in green is in the arcade" \\
        & ``A man strolls down a bustling city street under moonlight" \\
              & ``The man and a boy in green clothes are standing among the flowers by the lake" \\     
              & ``The man met a boy dressed in green at the foot of the tower" \\
        \bottomrule
    \end{tabular}
    \caption{Text prompts used in \cref{teaser}.}
\end{table*}

\section{Limitation and Discussion}
Although we have established an automated data curation framework, this paper primarily focuses on subject-driven generation. Our dataset currently contains limited editing and stylization data. While UNO is a unified and customizable framework with sufficient generalization capabilities, the types of synthetic data may somewhat restrict its abilities. In the future, we plan to expand our data types to further unlock UNO's potential and cover a broader range of tasks. 

\end{document}